%% file: main_full.tex
\pdfoutput=1

\documentclass[sigconf]{acmart}

\usepackage{color,soul}
\usepackage{lipsum}
\usepackage{hyperref}
\usepackage{enumitem}
\usepackage{caption}
\usepackage{subcaption}
\usepackage{cleveref}
\usepackage{xspace}
\usepackage{circledsteps}
\usepackage{adjustbox}
\usepackage[detect-weight=true]{siunitx}
\usepackage{multicol,multirow,booktabs,makecell,array}

\usepackage[linesnumbered]{algorithm2e}
\crefname{algocf}{algorithm}{algorithms}
\Crefname{algocf}{Algorithm}{Algorithms}
\usepackage{gensymb}

\usepackage{tikz,pgfplots,pgfplotstable}
\usetikzlibrary{patterns}
\pgfplotsset{compat=1.17}
\usepgfplotslibrary{groupplots,fillbetween,statistics}
\usetikzlibrary{external,decorations.pathreplacing,calligraphy,shapes,arrows,fit}
\tikzset{external/only named=true}
\usepackage{tabularx}
\usepackage[basecolor=blue!60,size=3mm]{vsrs}
\input{drawing}

\input{plot-macros}

\AtBeginDocument{%
 \providecommand\BibTeX{{%
 \normalfont B\kern-0.5em{\scshape i\kern-0.25em b}\kern-0.8em\TeX}}}

\copyrightyear{2024}
\acmYear{2024}
\setcopyright{rightsretained}
\acmConference[GECCO '24]{Genetic and Evolutionary Computation Conference}{July 14--18, 2024}{Melbourne, VIC, Australia}
\acmBooktitle{Genetic and Evolutionary Computation Conference (GECCO '24), July 14--18, 2024, Melbourne, VIC, Australia} 
\acmDOI{10.1145/3638529.3654011}
\acmISBN{979-8-4007-0494-9/24/07}

\begin{document}

\title{Neuron-centric Hebbian Learning}

\author{Andrea Ferigo}
\email{andrea.ferigo@unitn.it}
\affiliation{%
  \institution{University of Trento}
    \city{Trento}
  \country{Italy}
}

\author{Elia Cunegatti}
\email{elia.cunegatti@unitn.it}
\affiliation{%
  \institution{University of Trento}
    \city{Trento}
  \country{Italy}
}

\author{Giovanni Iacca}
\email{giovanni.iacca@unitn.it}
\affiliation{%
  \institution{University of Trento}
    \city{Trento}
  \country{Italy}
}

\renewcommand{\shortauthors}{Ferigo et al.}



\newcommand{\alghl}{\textcolor{cola1}{HL}\xspace}
\newcommand{\algnhl}{\textcolor{cola2}{NcHL}\xspace}
\newcommand{\algwnhl}{\textcolor{cola3}{WNcHL}\xspace}
\newcommand{\algdesc}{Neuron-centric Hebbian Learning\xspace}
\renewcommand{\vec}[1]{\boldsymbol{#1}}


\begin{abstract}
One of the most striking capabilities behind the learning mechanisms of the brain is the adaptation, through structural and functional plasticity, of its synapses. While synapses have the fundamental role of transmitting information across the brain, several studies show that it is the neuron activations that produce changes on synapses. Yet, most plasticity models devised for artificial Neural Networks (NNs), e.g., the ABCD rule, focus on synapses, rather than neurons, therefore optimizing synaptic-specific Hebbian parameters. This approach, however, increases the complexity of the optimization process since each synapse is associated to multiple Hebbian parameters. To overcome this limitation, we propose a novel plasticity model, called Neuron-centric Hebbian Learning (NcHL), where optimization focuses on neuron- rather than synaptic-specific Hebbian parameters. Compared to the ABCD rule, NcHL reduces the parameters from $5W$ to $5N$, being $W$ and $N$ the number of weights and neurons, and usually $N \ll W$. We also devise a ``weightless'' NcHL model, which requires less memory by approximating the weights based on a record of neuron activations. Our experiments on two robotic locomotion tasks reveal that NcHL performs comparably to the ABCD rule, despite using up to $\sim97$ times less parameters, thus allowing for scalable plasticity.
\end{abstract}


\begin{teaserfigure}
 \centering
 \includegraphics[width=0.7\textwidth]{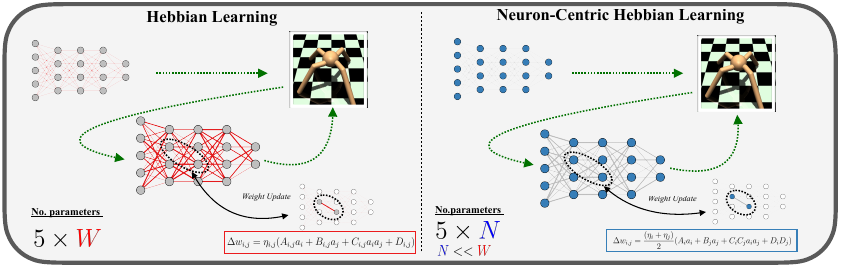}
 \vspace*{-0.2cm}
 \caption{Graphical representation of the traditional \alghl (left) and our \algdesc model (right). Regarding \alghl, when the network is initialized (first arrow) it starts a loop of observations and actions, in which \alghl updates the weights (black arrow) by using the equation shown in the red box (see also \Cref{eq:abcd}). As can be seen, the parameters of the Hebbian rule in \alghl are specific to each synapse, leading to a total of $5W$ parameters to optimize, with $W$ being the number of synapses. Alternatively, our \algnhl model (right) updates the weights (black arrow) by using the equation shown in the blue box (see also \Cref{eq:Nabcd}). In this case, the parameters of the Hebbian rule are specific to each neuron, hence resulting in a total of $5N$ parameters to optimize, with $N$ being the number of neurons. Assuming $N \ll W$, optimizing \algnhl models becomes notably less difficult than optimizing \alghl models, yet achieving comparable performance.}
 \Description{}
 \label{fig:teaser}
\end{teaserfigure}

\maketitle


\section{Introduction}
\label{sec:intro}
In the natural world, the number of neurons in the brain can vary by several orders of magnitude, from $302$ in the nematode worm \cite{nematode} to $70$ billion in Humans \cite{herculano2009human}. As known, these cells do not act independently, but it is their interaction that creates the complex network that we refer to as the brain. This complex system shows the capability to solve complex, heterogeneous tasks and adapt to new and unforeseen situations. This kind of adaptation is currently still limited in modern Artificial Intelligence (AI) systems.

One of the most important mechanisms that directly affects the adaption capability of the brain is the \emph{synaptogenesis}, which regulates the growth and pruning of the synapses over the life span of an organism. For example, during the lifetime of a person, there is a growing phase where the number of synapses reaches a staggering number of $4$ quadrillion synapses, eventually decreasing during adulthood and stabilizing at around $500$ trillion \cite{synN,ZUKERMAN20119}. Recent attempts tried to replicate this process in artificial Neural Networks (NNs) \cite{ferigo2023sbm}, or characterize the similarity between natural and artificial NNs \cite{you2020graph}.

The second important mechanism behind brain adaptation is its way to \emph{change the strength} of its synapses, based on the frequency with which they are used. This process, originally described by Hebb in \cite{hebb2005organization}, is at the base of the concept of \emph{plasticity}. Differently from the well-known (but, somehow less biologically plausible) gradient-based backpropagation, which is a supervised learning mechanism based on backward propagation (from outputs to inputs) of errors over a NN, Hebbian Learning (\alghl\footnote{In this paper, we will use \alghl to refer to the traditional, Synaptic-Centric Hebbian Learning model (including the ABCD rule model), to distinguish it from our proposed \algdesc model, referred to as \algnhl.}) is a biologically plausible mechanism that allows learning by only propagating information \emph{forward} (from inputs to outputs) over the network.

Different \alghl models have been proposed in the literature, addressing various tasks such as computer vision classification \cite{softHebb}, test-time adaptation \cite{tang2023neuro}, and Reinforcement Learning (RL) for robotics \cite{el2011reinforcement,10.3389/fnbot.2019.00081}, where, e.g., a robot needs to learn how to perform locomotion \cite{ferigo2022evolving} or maze navigation \cite{yaman2021evolving}, or adapt to unforeseen situations, such as damages in its morphology \cite{najarro2020meta}. 
However, the existing \alghl methods focus on the synapses, i.e., each connection has its own Hebbian rule to control the weight update, with its corresponding parameters. Hence, from an optimization perspective, the number of parameters to optimize in a Hebbian model increases linearly with the number of synapses in the network. This makes the training of \alghl models more computationally demanding.
 
In this work, we address this issue from a biologically plausible perspective. Leveraging on the intuition that, in \alghl, it is the activation of the neurons that produces a change of the synapses, we depart from the traditional, \emph{Synaptic-centered} \alghl paradigm to propose a novel, \emph{\algdesc} model (in short, \algnhl) where the parameters of the Hebbian update rule are specific to neurons rather than synapses. This leads to a significant decrease in the number of parameters of the models, as the number of neurons in an NN can be orders of magnitude smaller than the number of synapses. Moreover, even though in our model the parameters of the update rule are specific to each neuron rather than each synapse, the model still allows for specialization of the synaptic updates, thanks to the effect of the parameters associated to the pre- and post-synaptic neurons connected to each synapse.

As a second contribution of this work, we then reduce the \algnhl model to what we call a ``weightless'' version of it (\algwnhl). This model does not use the weights to calculate the activations, but each neuron stores a record of previous activations and uses these values to calculate the weights without explicitly storing them.

We tested our method on two simulated robotic locomotion tasks, one involving Voxel-based Soft Robots (VSR) \cite{medvet20202d}, and one involving a quadruped robotic task from the MuJoCo suite \cite{todorov_mujoco_2012}, simulated in the PyBullet physics engine \cite{coumans2021}. For these robots, we used Feed-Forward NNs of different sizes, moving from the smallest network with only $13$ neurons and $30$ synapses, to the biggest one with $420$ neurons and $40960$ synapses. 

In a first set of experiments, our results showed that the \algnhl model reaches comparable performance as the traditional \alghl model for all the tasks and NNs tested. However, the \algnhl model needs to optimize up to $\sim97$ times less parameters. In a second set of experiments, we then tested the \algwnhl model and observed that optimizing its Hebbian parameters requires $\sim99.7\%$ less memory than \algnhl, while losing only between $\sim17\%$ and $\sim37\%$ of the performance. Finally, we compared the behavior of the \alghl and \algnhl models, showing that the two models tend to explore the behavioral space following similar trajectories.
 
The rest of this paper is organized as follows. \Cref{sec:rw} introduces the background and summarizes the related works. \Cref{sec:methodology} describes the methods. \Cref{sec:experiments} presents the experimental setup and the numerical results, followed by the conclusions in \Cref{sec:conclusions}.


\section{Background and related work}
\label{sec:rw}

Plasticity refers to the capability of the brain to adapt internally, in order to learn to deal with new situations during the agent's lifetime, a fundamental property for both natural \cite{baldwin1896new} and artificial learning systems \cite{hinton1996learning,li2023evaluation}.

In living organisms, plasticity derives from the capability of the brain to create, change, or destroy, its synapses. Creating and destroying synapses are seen as a form of \emph{structural plasticity} \cite{shaw2001toward}, as they lead to a direct modification of the structure of the brain \cite{zilles1992neuronal,patten2015benefits}. Changing (i.e., strengthening or weakening) the existing synapses, instead, is seen as a form of \emph{functional plasticity}, as it modifies the processing function of the brain. More specifically, the capability of the brain to change the strength of synapses, based on the activations of the neurons they connect, is known as Hebbian Learning (\alghl) \cite{brown1990hebbian, hebb2005organization}, which postulates that neurons that often fire together will eventually strengthen their connection, while neurons that do not influence each other will tend to weaken it.

\subsubsection*{\textbf{Structural plasticity}}
The concepts above have been extensively studied in artificial NNs. In particular, structural plasticity has been translated into various approaches to \emph{pruning} and \emph{network development}. The former consists of selecting some synapses to be removed from a dense (i.e., fully connected) NN, without affecting its performance. In this case, the overall size of the network (in terms of number of neurons) does not change, while the number of parameters (i.e., weights) can be greatly decreased \cite{Hoefler,gale2019sparsity,mocanu2018scalable}. The recent literature has thoroughly studied the most relevant characteristics of sparse NNs, both for Feed-Forward \cite{stier2019structural,liu2021topological} and Convolutional NNs \cite{pal2022study,cunegatti2023peeking}. Another work \cite{you2020graph} observed structural similarity between high-performing sparse NNs and biological systems \cite{bassett2006small}.

Network development, instead, takes a different approach. It builds the NN from an initial (typically small, random) structure, to evolve it into a more complex, possibly more efficient one \cite{najarro2023selfassembling,ferigo2023sbm}. In this case, both neurons and synapses are iteratively added to the network, to improve its capacity to learn on the given task. 

\subsubsection*{\textbf{Functional plasticity (Hebbian Learning)}}
Functional plasticity in artificial NNs refers to the capability of the network to change its weights during the execution of a certain task. In this case, \alghl specifically refers to how weights are updated based on a given \emph{Hebbian rule}, which determines the perturbation to be applied to the weights at each step of the task execution. These rules can be either discovered through an optimization process, such as Genetic Programming \cite{jordan2021evolving} or autoML \cite{autoMLZ} (as done in \cite{10341979}) or, most frequently, handcrafted before the task execution.
In this latter case, the Hebbian rules are based on a fixed parametric formulation, and only the parameters of such rules are optimized.

It is worth noticing that, differently from gradient-based backpropagation, Hebbian learning does not require any backward pass. In fact, the NNs weights are updated following the selected Hebbian rules after at \emph{each forward step}.
To better evaluate the Hebbian parameters, the performance of an agent is computed \emph{globally} w.r.t. the cumulative reward over the task, rather than \emph{locally}, i.e., as the performance at each forward step. During this process, the Hebbian rule is fixed, and its parameters are typically updated through evolutionary optimization.

Different Hebbian rules can be found in the literature \cite{soltoggio2018born}. One of the most established models is the so-called \emph{ABCD rule}, which has been proven effective in different tasks, achieving comparable or better performance w.r.t. non-plastic NNs \cite{mattiussi2007analog, niv2001evolution,najarro2020meta,ferigo2022evolving}. 
In this model, each weight $w_{i,j}$ is updated as:
\begin{align}
 w_{i,j} = w_{i,j} + \Delta w_{i,j}
 \label{eq:updatew}
\end{align}
where $\Delta w_{i,j}$ is a function $h(a_i, a_j)$ calculated as follows:
\begin{align}
 \Delta w_{i,j} = \eta_{i,j} \left(A_{i,j} a_i + B_{i,j} a_j + C_{i,j} a_i a_j + D_{i,j}\right)
 \label{eq:abcd}
\end{align}
where $\eta_{i,j}$ is the learning rate, $A_{i,j}$ modulates the activation of the pre-synaptic neuron ($a_i$), $B_{i,j}$ modules the activation of the post-synaptic neuron ($a_j$), $C_{i,j}$ modulates the mutual interaction of the two activations ($a_i, a_j$) and $D_{i,j}$ is a bias. 
Namely, the $A$, $B$, $C$, and $D$ parameters can be seen as the weights of a non-linear combination (scaled by the learning rate $\eta$) of the pre and post-synaptic activation values and a unitary bias. 
If one of the four parameters is greater than the others (in absolute value), it means that the corresponding input value is more relevant to the output (i.e., $\Delta w_{i,j}$) than the other inputs. 
For instance, if the bias (associated with $D$) turns out to be the only relevant (i.e., non-null) input value, $\Delta w_{i,j}$ will not be affected by the activation values but rather will have a fixed increase. We suppose that this could happen in cases where the previous forward steps do not influence the next one (i.e., in a classification task where subsequent samples are uncorrelated, hence the forward step does not depend on the previous ones).

In principle, it could be possible that one single set of parameters ($A, B, C, D$, plus, possibly, $\eta$) is for all the synapses in the network (i.e., 
$A_{i,j}=A; B_{i,j}=B; C_{i,j}=C; D_{i,j}=D; \eta_{i,j}=\eta~\forall~{i,j}$). However, this approach would not allow for specialization of the weight updates across the different parts of the network, hence resulting in ineffective learning \cite{najarro2020meta, ferigo2022evolving}. Hence, the majority of works dealing with the ABCD rule in practice optimize a set of $5$ parameters \emph{for each synapse}. The drawback of this approach is, obviously, that the total number of parameters to optimize increases from $W$ (i.e., the number of weights) as in traditional NN training, to $5W$, since each synapse is associated to a set of $5$ different parameters. This is a common aspect across various plasticity models in the literature \cite{najarro2020meta, ferigo2022evolving,soltoggio2007evolving}, with some exceptions such as \cite{yaman2021evolving} where, in order to reduce the number of \alghl parameters, $\Delta w_{i,j}$ is set to 
$\eta_{i,j} m_{i,j} a_i a_j$, 
where $\eta_{i,j}$ is a learning rate and $m_{i,j}$ is a neuro-modulatory signal, with both $\eta_{i,j}$ and $m_{i,j}$ being optimized separately for each synapse.
Another method from the literature to reduce the number of parameters is the merging procedure described in \cite{Pedersen2021Merging}: during the evolution, this algorithm merges similar rules using k-means and then starts a new optimization with the merged model. Repeating this procedure, the number of parameters is halved until a desired value. However, this operation requires restarting the evolutionary process increasing the number of iterations, and requires a clustering method.

\subsubsection*{\textbf{Novelty of the proposed approach}}
In this work, we shift the focus of the Hebbian update rules from synapses to the neurons, to better mimic the behavior of biological brains \cite{brown1990hebbian}, where neurons, rather than synapses, are the elements of specialization. This paradigm shift allows us to greatly reduce the number of parameters to optimize, as the number of neurons ($N$) in an NN is, apart from the case of heavily sparsified networks, several times lower than the number of synapses ($W$). As we will see in the next section, this is the main advantage of our proposal.


\section{Methods}
\label{sec:methodology}
As seen in the previous section, given an NN with $W$ synapses, the traditional ABCD rule optimizes $5W$ parameters. Here, we first present our \algnhl model that reduces the number of parameters to optimize to just $5N$, where $N$ is the number of neurons, rather than synapses in the NN (as said, we can assume that $N \ll W$). Then, we discuss the formalization of \algnhl, as well as its ``weightless'' variant (\algwnhl), which is based on an approximation based on the previous neuron activations and is specifically designed to avoid storing weight parameters at all. Finally, we present the Evolution Strategies algorithms that we use to optimize our models~\footnote{Our codebase is publicly available at \url{https://github.com/ndr09/NCHL}.}.

\subsection{\algdesc}
\label{sec:neuronHebbian}
The main idea behind our proposed \algnhl model is to specialize the weight update at the level of neuron rather than synapse, still preserving the inherent independence of the traditional ABCD rule w.r.t. the reward of the given task. In other words, the weight update is solely regulated by the activations of the pre- and post-synaptic neurons $i$ and $j$. However, differently from the traditional ABCD rule, see \Cref{eq:abcd}, where the parameters $A_{i,j}$, $B_{i,j}$, $C_{i,j}$, $D_{i,j}$, and $\eta_{i,j}$ can be specific to each synapse, in \algnhl the Hebbian rule parameters are specific \emph{to each neuron}, i.e., each neuron has a different set of $5$ parameters, and each weight $w_{i,j}$ is updated according to the following update rule: $h(a_i, a_j)$:
\begin{align}
 \Delta w_{i,j} = \frac{(\eta_i+\eta_j)}{2} (A_i a_i + B_j 
 a_j + C_i C_j a_i a_j + D_i D_j)
 \label{eq:Nabcd}
\end{align}
where: the learning rate is calculated as $\frac{(\eta_i+\eta_j)}{2}$, i.e., we consider the average learning rate of the pre- and post-synaptic neurons; $A_i$ and $B_i$ are associated, respectively, to the pre-synaptic neuron $i$ and the post-synaptic neuron $j$, and modulate their corresponding activations; the product $C_i C_j$ modulates the co-contribution of the activations of the pre- and post-synaptic neurons; the product $D_i D_j$ is calculated by multiplying the bias associated to the pre- and post-synaptic neurons.

Even though the Hebbian rule update shown in \Cref{eq:Nabcd} turns out to be more complex compared to the traditional ABCD rule, it is important to note that the total number of parameters in the whole network greatly decreases, as we will demonstrate in \Cref{sec:experiments}. In the traditional model, the total number of parameters to optimize increases linearly with the number of synapses. Instead, in \algnhl it increases linearly with the number of neurons, hence resulting in a smaller number of parameters to optimize.

\subsection{\algdesc ``weightless''}
With the \algnhl model, we reduce the number of parameters needed to \emph{store} from $6W$ for the traditional ABCD model ($W$ weights plus $5W$ Hebbian rule parameters) to $W+5N$ ($W$ weights plus $5N$ Hebbian rule parameters). Here, we further propose a new Hebbian model aiming to further reduce the memory needed to store the model parameters. In particular, rather than storing the weights in memory, we approximate them using the \algnhl update rule. Hence, we call this this model ``weightless'' \algnhl (in short, \algwnhl).

To explain the rationale behind this model, we start with the following considerations. Let us assume we have two neurons $i$ (pre-synaptic) and $j$ (post-synaptic) with their respective activations $a_i$ and $a_j$, which are linked by $w_{i,j}$. Now, if we assume that the initial value of $w_{i,j}$, $w_{i,j}^0$, is $0$\footnote{Since reducing the memory needed to store the model parameters is the main requirement of the proposed \algwnhl method, this assumption allows us to avoid storing $W$ additional parameters for the initial values of the weights.}, we can express the value of the activation of the post-synaptic neuron $j$ at step $T$ as $a_j^T = f(w_{i,j}^T a_i^T)$, where $f(\cdot)$ is the activation function and $w_{i,j}^T = \sum_{t=0}^T h(a_i^t,a_j^t)$, where $h(\cdot)$ is the Hebbian update rule expressed in \Cref{eq:Nabcd}\footnote{Note that, in principle, this can be applied also to the traditional ABCD rule in \Cref{eq:abcd}. However, in this case, the reduction in terms of the number of parameters would be only $W$.}. 
In other words, by computing $h(a_i^t,a_j^t)=\Delta_{i,j}$ for each step $t$, we can find the value of $w_{i,j}$ at step $T$. This means that, rather than storing $w_{i,j}$, it is possible to calculate it only based on the history of pre-synaptic and post-synaptic activations over the various steps until $T$.

However, storing the activations for all steps will lead to using more memory than what is needed if we store the weights. Hence, rather than considering the whole history of steps until $T$, we store only a subset of the last activations, aiming to reduce memory consumption. We call this subset the \emph{memory window}, whose size is $M_w$. Therefore, we approximate the weight values as follows:
\begin{align}
 w_{i,j}^{M_w} = \sum\nolimits_{t=0}^{M_w} h(a_i^t,a_j^t).
 \label{eq:weightless}
\end{align}
In other words, to calculate $w_{i,j}^{M_w}$, it is enough to store the activations of the neurons over $M_w$ steps in the memory window.

\subsection{Evolution Strategies}
To optimize the Hebbian rules, we employ two Evolution Strategies (ES). The first one, provided in \cite{medvet2022jgea}, creates a new population of $m$ solutions in two steps: firstly, it selects the $n$ best solutions and calculates the vector $\vec{g}_{mean}$ as the mean of these solutions, namely $\vec{g}_{mean} = 1/m\sum_{i=0}^{m}{\vec{g}_i}$. Then, it generates $m-1$ new solutions by adding to each variable of $\vec{g}_{mean}$ a random value sampled from $\mathcal{N}(0,\sigma)$.
Finally, the algorithm composes the new population by combining the $m-1$ new solutions with the best solution from the previous generation. We refer to this algorithm as ES$_1$.

The second algorithm, from \cite{najarro2020meta}, generates the new population by updating the solutions based on their fitness. 
For each solution $\vec{g}$ in a population of size $m$, a random perturbation $\mathcal{N}(0, \sigma)$ is added to each variable, $\vec{g}'= \vec{g}+N(0,\sigma)$. 
Then, based on the fitness of the solution, each parameter of $\vec{g'}$ is updated as follows: $\vec{g''} = {lr}/{(m\sigma)}+\sum_{i=0}^{m}{\vec{g}_{i}'F(\vec{g}_{i}')}$. 
Note that $\sigma$ and $lr$ decay over the generation by a factor $lr_{decay}$ and $\sigma_{decay}$. We refer to this algorithm as ES$_2$.


\section{Experiments}
\label{sec:experiments}

We performed a suite of experiments aimed at answering the following research questions:
\begin{enumerate}[label=RQ\arabic*,leftmargin=*]
 \item \label{item:rq-1} How does \algnhl perform w.r.t. the traditional \alghl model based on the ABCD rule?
 \item \label{item:rq-2} Does the \algwnhl model achieve similar performance w.r.t. \algnhl? 
 \item \label{item:rq-3} Since \algnhl and \alghl models optimize the Hebbian rules parameters respectively w.r.t. neurons and synapses, are the best solutions found by the two models somehow behaviorally similar, or different?
\end{enumerate}

\subsection{Experimental setup}

\subsubsection*{\textbf{Tasks}}
We tested our model on two simulated locomotion tasks, namely \Circled{1} one where we employed Voxel-based Soft Robots (VSR) \cite{medvet20202d,medvet2020design}, and \Circled{2} one where we employed the RL task Ant \cite{schulman2018highdimensional} from the MuJoCo suite \cite{todorov_mujoco_2012}. 
The first task has been selected for being somehow easier from an optimization perspective. Hence, we tested it on smaller NNs. On the other hand, the second task, which is well-established for its complexity, has been selected to test the scalability of our approach, as this task requires much larger NNs in order to be solved effectively.
For both tasks, the goal is to make the agent walk the farthest, so the fitness is the distance covered by the robot from the beginning to the end of each task episode.

The first task \Circled{1} is characterized by robots whose body is composed of voxels (i.e., soft structures that can shrink or expand based on a control signal). 
Each voxel can sense various information, namely: (1) the velocity of the voxel (along the $x$ and $y$ axes), (2) the ground contact, (3) the area ratio (namely, the compression or expansion of the voxel area w.r.t. its rest surface), and (4) the distance from an obstacle or the terrain (through a Lidar sensor). The locomotion happens in an uneven terrain composed of hills with different slopes, lasting $3600$ steps. 
For this task we relied on the 2D VSR simulator implemented in \cite{medvet20202d,medvet2020design}.
We considered two kinds of morphology shapes, namely a $7\times1$ shape, a.k.a. a \emph{worm}, and $4\times3$ shape with a $2\times1$ ``hole'' in the central bottom part, a.k.a. a \emph{biped}. We used these shapes as they have been successfully employed in previous works regarding this kind of VSR \cite{ferigo2022evolving,ferigo2022entanglement,NADIZAR2023110610}.
We also varied the sensors equipped on these robots, defining three sensory configurations that we refer to respectively as \emph{low}, \emph{medium}, and \emph{high}, depending on the number of sensors present in each voxel. In particular, in the \emph{low} configuration, there are only area ratio sensors, the \emph{medium} configuration uses also the velocity and ground contact sensors, while the \emph{high} configuration uses all four kinds of sensors. The total number of sensor inputs for each configuration is $3$, $28$, and $31$ for the worm and $8$, $20$, and $29$ for the biped, respectively for the \emph{low}, \emph{medium}, and \emph{high} configurations. More details on these three configurations can be found in \cite{ferigo2022evolving}. In total, we tested six VSR configurations (two shapes, each one with three sensory configurations).

In the second task \Circled{2}, a 3D quadruped robot, called Ant, has to move as far as possible, for a total of $1000$ steps, on a flat terrain \cite{schulman2018highdimensional}.
In this task, the agent has to control $8$ actuated joints (two per leg) based on a set of $28$ inputs representing the velocity and position of the different parts of the robot. For this task, we used the implementation available in the PyBullet physics engine \cite{coumans2021}.

\subsubsection*{\textbf{Neural network architectures}}
We used a set of different Feed-Forward NNs for each task. We used the $tanh$ activation function.


\Circled{1} Regarding the VSR, as discussed before, the number of inputs depends on the sensory configuration and shape of the VSR. As for the number of outputs, it depends on the number of voxels present in the shape, thus $7$ for the worm and $10$ for the biped. Following \cite{ferigo2022evolving}, all the NNs used for this task have been configured with a single hidden layer with the same size as the input, so in total we have six NN configurations, i.e., one per VSR configuration.

\Circled{2} For the Ant task, the sensory configuration and structure of the robot are fixed. Hence, we tested two different NN configurations with two hidden layers. The first one, which we refer to as \emph{medium} has two layers of sizes $128$ and $64$, while the second, which we refer to as \emph{high}, has two layers of sizes $256$ and $128$.

\subsubsection*{\textbf{Evolution Strategies}}
For the eight networks described above (six for VSR and two for Ant), we optimized two Hebbian models, namely the traditional ABCD rule (that we refer to as \alghl), and our proposed \algnhl model. 
We report in \Cref{tab:task_p} all the parameters (size of input, hidden and output layers) of the NNs used for both tasks, including the different number of parameters to optimize for each network configuration using either \alghl or \algnhl, along with the ratio of the number of parameters between \alghl and \algnhl. Note that, in the VSR task the $\eta$ value is fixed at $\eta=0.1$ (as in previous works on VSR \cite{ferigo2022evolving,ferigo2022entanglement,NADIZAR2023110610}), while for the Ant task, it is optimized together with the other Hebbian parameters. Also, we should note that, for both ES$_1$ and ES$_2$, we used direct representation, i.e., the algorithms directly evolved the Hebbian parameters of the two models.

For the experiments for RQ1, we decided to use ES$_1$ for the VSR task, as this algorithm is provided with the VSR simulator \cite{medvet2022jgea}, while we used ES$_2$ for the Ant task, as its implementation is already provided with the implementation of the task \cite{najarro2020meta}. 
For the experiments for RQ2, instead, we preferred to use ES$_1$ for the Ant task, because of constraints on the computational resources.

ES$_1$ is, in fact, configured to be less computationally expensive, with a population size of $40$, $500$ generations, and $\sigma=0.35$. ES$_2$, on the other end, is configured to be more computationally demanding, with a population size of $500$, $500$ generations, $\sigma=0.1$, learning rate $lr =0.2$, $\sigma_{decay}= 0.999$, and $lr_{decay}=0.995$.

Note that, for the experiments on RQ1 and RQ3, we uniformly initialized the weights of the networks in the range $[-0.1, 0.1]$, while for RQ2 all weights are set to $0$.
Moreover, we remark that for each experiment and setting in RQ1 and RQ2, we collected the results from $10$ independent runs of the corresponding ES.

\subsubsection*{\textbf{Random baselines}}
In both tasks, we also tested two random baselines, which we refer to as \alghl-Random and \algnhl-Random. 
\alghl-Random solutions are created by associating, independently for each synapse $(i,j)$, a Hebbian rule with its parameters ($A_{i,j}$, $B_{i,j}$, $C_{i,j}$, $D_{i,j}$, $\eta_{i,j}$) uniformly sampled in $[-1,1]$, see \Cref{eq:abcd}.
Similarly, \algnhl-Random solutions are created by associating, independently for each neuron $i$, a Hebbian rule with its parameters ($A_i$, $B_i$, $C_i$, $D_i$, $\eta_i$) uniformly sampled in $[-1,1]$, see \Cref{eq:Nabcd}.
Since, for the experiments conducted to address RQ1 and RQ2, the performances of these random baselines are close to zero for both tasks, we do not present those results in the analysis of those research questions. Instead, we report the behavior of the random baselines only in the analysis conducted for RQ3.

\begin{table}[ht!]
 \vspace{-0.3cm}
 \centering
 \caption{Network description and number of parameters to optimize for the different settings we tested. In the VSR task, we use an NN with a single hidden layer whose size is equal to the input size. Here, the \algnhl model requires optimizing up to $\sim$21 times less parameters than \alghl, depending on the network size. In the Ant task, where we used a 2-hidden layer NN, the advantage of \algnhl in terms of the number of parameters to optimize w.r.t. \alghl model increases: \algnhl requires up to $\sim$97 times less parameters than \alghl.}
 \label{tab:task_p}
 \vspace{-0.1cm}
 \resizebox{\columnwidth}{!}{%
 \begin{tabular}{
 c
 c
 c
 r
 r
 r
 r
 r
 r
 }
 \toprule
 \multirow{2}[2]{*}{\textbf{Task}} & \multirow{2}[2]{*}{\textbf{Shape}} & \multirow{2}[2]{*}{\parbox{2cm}{\centering\textbf{NN}\\\textbf{Configuration}}} & \multicolumn{3}{c}{\textbf{Layer size}} & \multicolumn{3}{c}{\textbf{No. of parameters}}\\
 \cmidrule(r){4-6} \cmidrule(l){7-9}
 & & & \multicolumn{1}{r}{\textbf{Input}} & \multicolumn{1}{r}{\textbf{Hidden}} & \multicolumn{1}{r}{\textbf{Output}} & \multicolumn{1}{r}{\textbf{\alghl}} & \multicolumn{1}{r}{\textbf{\algnhl}} & \multicolumn{1}{r}{\textbf{Ratio}} \\
 \midrule
 \multirow{6}{*}{VSR} & \multirow{3}{*}{Biped} & Low & 8 & 8 & 10 & 576 & 88 & 6.54\\
 & & Medium & 20 & 20 & 10 & 2400 & 160 & 15.00\\
 & & High & 29 & 29 & 10 & 4524 & 214 & 21.14\\
 \cmidrule{2-9}
 & \multirow{3}{*}{Worm} & Low & 3 & 3 & 7 & 120 & 46 & 2.60\\
 & & Medium & 28 & 28 & 7 & 3920 & 196 & 20.00\\
 & & High & 31 & 31 & 7 & 4712 & 214 & 22.01\\
 \midrule
 \multirow{2}{*}{Ant} & \multirow{2}{*}{-} & Medium & 28 & {128, 64} & 8 & 61440 & 1140 & 53.89\\
 & & High & 28 & {256, 128} & 8 & 204800 & 2100 & 97.52\\ 
 \bottomrule
 \end{tabular}
 }
 \vspace{-0.5cm}
\end{table}


\subsection{Experimental results}

We now present the results of the experiments conducted to answer the three RQs discussed above.

Before going into the details of the three RQs, it is worth highlighting once again that the proposed \algnhl model significantly reduces the number of parameters to optimize w.r.t. the traditional \alghl model. This trend is clearly visible in \Cref{tab:task_p}, see in particular the last three columns, where we summarize the number of parameters to optimize in each case. We can observe that the total number of parameters to optimize with the proposed \algnhl model is one or two orders of magnitude smaller than the \alghl model. In the best case (the ``high'' configuration on the Ant task) \algnhl requires $\sim97$ times less parameters than \algnhl. 

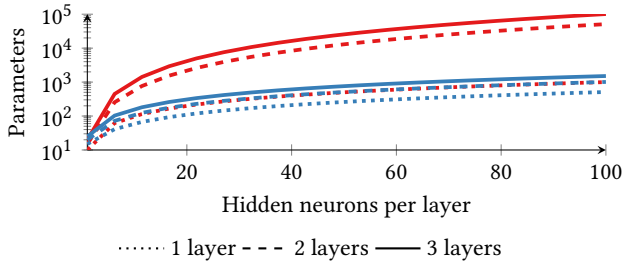
\begin{figure}[ht]
 \centering
 \begin{tikzpicture}
 \begin{semilogyaxis}[
 axis lines = left,
 xlabel = {Hidden neurons per layer},
 ylabel = {Parameters},
 width=\columnwidth,
 height=0.4\columnwidth,
 legend columns=2, 
 legend entries={\alghl, \algnhl},
 legend to name=legendComparison,
 legend style={draw=none},
 ytick={1e0,1e1,1e2,1e3,1e4,1e5}
 ]
 \addlegendimage{solid,cola1}
 \addlegendimage{solid,cola2}
 \addplot [domain=1:100, samples=20, dotted,color=cola1, line width=0.5mm]{5*(2*x)};
 \addplot [domain=1:100, samples=20, dashed, color=cola1, line width=0.5mm]{5*(2*x+x*x)};
 \addplot [domain=1:100, samples=20, color=cola1, line width=0.5mm]{5*(2*x+2*x*x)};
 \addplot [domain=1:100, samples=20, dotted, color=cola2, line width=0.5mm] {5*(x+2)};
 \addplot [domain=1:100, samples=20, dashed, color=cola2, line width=0.5mm] {5*(2*x+2)};
 \addplot [domain=1:100, samples=20, color=cola2, line width=0.5mm] {5*(3*x+2)};
 \end{semilogyaxis}
 \end{tikzpicture}\\
 \tikzexternaldisable\pgfplotslegendfromname{la} \tikzexternalenable
 \vspace*{-0.4cm}
 \caption{Number of parameters to optimize with \alghl and \algnhl in Feed-Forward NN (toy example, 1 input and 1 output) when varying the no. of hidden layers and neurons therein.}
 \Description{}
 \label{fig:weightsComparison}
 \vspace{-0.4cm}
\end{figure}

To further demonstrate this advantage of our model, \Cref{fig:weightsComparison} displays the number of parameters needed (on a toy example of a task with one input and one output) by \alghl and \algnhl when increasing both the number of hidden layers and neurons therein (note the logarithmic scale on the y-axis). We can clearly observe that \algnhl requires a substantially smaller number of parameters w.r.t. a \alghl model with the same number of hidden layers.

While this mechanism can reduce the size of the search space, it could also lead to subpar performances as it might be harder for the evolution to find solutions with enough behavioral complexity to solve the task at hand. Therefore, we are interested in studying the performance that our proposed \algnhl model can achieve w.r.t. the traditional \alghl model. This is the focus of our RQ1.

\subsubsection*{\textbf{RQ1: \algnhl vs \alghl performance}}

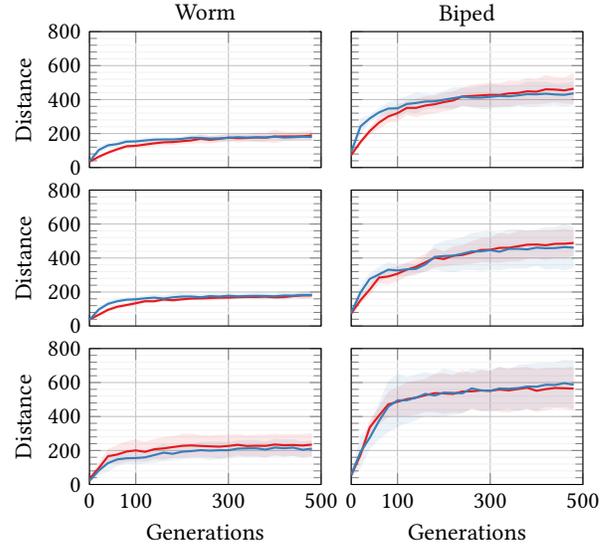
\begin{figure}[ht!]
 \centering
 \begin{tikzpicture}
 \begin{groupplot}[
 width=0.55\linewidth,
 height=0.40\linewidth,
 grid=both,
 grid style={line width=.1pt, draw=gray!10},
 major grid style={line width=.2pt,draw=gray!50},
 minor tick num=4,
 title style={anchor=north, yshift=2ex},
 group style={
 group size=2 by 3,
 horizontal sep=4mm,
 vertical sep=3mm,
 xticklabels at=edge bottom,
 yticklabels at=edge left
 },
 scaled x ticks = false,
 x tick label style={
 /pgf/number format/.cd,
 fixed,
 fixed zerofill,
 int detect,
 1000 sep={},
 precision=3
 },
 ylabel style={
 align=center
 },
 every axis plot/.append style={thick},
 ymin=0,ymax=800,
 xmin=0,xmax=500,
 xtick={0,100,300,500}
 ]
 \nextgroupplot[
 legend columns=2,
 legend entries={\alghl, \algnhl},
 legend style={draw=none},
 legend to name=legendRQ1lines,
 ylabel={Distance},
 title={Worm}
 ]
 \linewitherror{data/rq1/ft/hwh.txt}{i}{m}{s}{cola1}
 \linewitherror{data/rq1/ft/nwh.txt}{i}{m}{s}{cola2}
 
 \nextgroupplot[
 title={Biped}
 ]
 \linewitherror{data/rq1/ft/hbh.txt}{i}{m}{s}{cola1}
 \linewitherror{data/rq1/ft/nbh.txt}{i}{m}{s}{cola2}
 
 \nextgroupplot[
 ylabel={Distance}
 ]
 \linewitherror{data/rq1/ft/hwm.txt}{i}{m}{s}{cola1}
 \linewitherror{data/rq1/ft/nwm.txt}{i}{m}{s}{cola2}
 
 \nextgroupplot[
 ]
 \linewitherror{data/rq1/ft/hbm.txt}{i}{m}{s}{cola1}
 \linewitherror{data/rq1/ft/nbm.txt}{i}{m}{s}{cola2}
 
 \nextgroupplot[
 ylabel={Distance},
 xlabel={Generations}
 ]
 \linewitherror{data/rq1/ft/hwl.txt}{i}{m}{s}{cola1}
 \linewitherror{data/rq1/ft/nwl.txt}{i}{m}{s}{cola2}
 
 \nextgroupplot[
 xlabel={Generations}
 ]
 \linewitherror{data/rq1/ft/hbl.txt}{i}{m}{s}{cola1}
 \linewitherror{data/rq1/ft/nbl.txt}{i}{m}{s}{cola2}
 
 \end{groupplot}
 \end{tikzpicture}
 \vspace*{-0.2cm}
 \caption{RQ1: Distance trend of the best solution (avg. $\pm$ std. dev. across 10 runs) found at each generation by \alghl and \algnhl in the various configurations of the VSR task. From top to bottom, the results refer to the High, Medium, and Low sensory configurations.}
 \Description{}
 \label{fig:fitness-evolution-vsr}
 \vspace{-0.1cm}
\end{figure}

Firstly, we analyzed the results of the VSR task \Circled{1}. 
In \Cref{fig:fitness-evolution-vsr}, we report the fitness trends for the different shapes and sensory configurations.
We can observe that the \algnhl and \alghl models show a very similar trend for both the worm and the biped shapes and for all three sensory configurations.
In \Cref{fig:rq1-bp-vsr}, we show the distribution of the distance achieved by the best solution found in each independent run. 
While, in some cases, the distribution seems slightly different, conducting the TOST test \cite{Vallat2018} with a bound of $10\%$ on the average fitness returned a $p$-value greater than $0.05$ for all the pairwise comparisons, indicating that the hypothesis of statistical equivalence cannot be rejected.

\begin{figure}
 \centering
 \begin{tikzpicture}
 \begin{groupplot}[
 boxplot,
 boxplot/draw direction=y,
 width=1\linewidth,
 height=0.45\linewidth,
 group style={
 group size=1 by 1,
 horizontal sep=1mm,
 vertical sep=1.5mm,
 xticklabels at=edge bottom,
 yticklabels at=edge left,
 },
 ymin=500,ymax=1800,
 legend cell align={left},
 ymajorgrids=true,
 yminorgrids=true,
 xtick={1.5, 3.5, 5.5, 7.5},
 xticklabels={{Ant\\Medium}, {Ant\\High}},
 ticklabel style={font=\footnotesize},
 grid style={line width=.1pt, draw=gray!10},
 major grid style={line width=.15pt, draw=gray!50},
 minor y tick num=4,
 title style={anchor=north, yshift=2ex},
 xmajorticks=true,
 xminorticks=true
 ]
 \nextgroupplot[
 align=center,
 legend columns=2,
 legend entries={\alghl, \algnhl},
 legend style={draw=none},
 ylabel={Distance},
 legend to name=legendRQ1box-ant
 ]
 \addlegendimage{area legend,color=cola1,fill}
 \addlegendimage{area legend,color=cola2,fill}
 
 \addplot[black,fill=cola1] table[y=h] {data/rq1/bp/am.txt};
 \addplot[black,fill=cola2] table[y=n] {data/rq1/bp/am.txt};
 \pvalue{1}{2}{1450}{50}{$0.26$}
 \draw [dashed] (2.5,\pgfkeysvalueof{/pgfplots/ymin}) -- (2.5,\pgfkeysvalueof{/pgfplots/ymax});
 \addplot[black,fill=cola1] table[y=h] {data/rq1/bp/ah.txt};
 \addplot[black,fill=cola2] table[y=n] {data/rq1/bp/ah.txt};
 \pvalue{3}{4}{1600}{50}{$0.66$}
 
 \end{groupplot}
 \end{tikzpicture}
 \vspace*{-0.2cm}
 \caption{RQ1: Distribution of the average distance in 100 rollouts of each best solution found in each of the 10 runs of \alghl and \algnhl in the two configurations of the Ant task.}
 \Description{}
 \label{fig:rq1-bp-ant}
 \vspace{-0.5cm}
\end{figure}
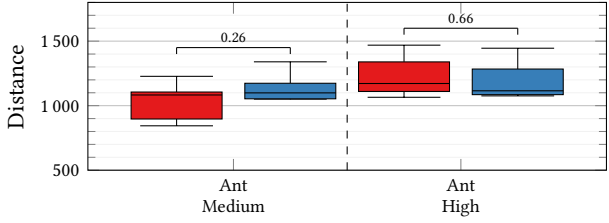

We then moved to analyze the results from the Ant task \Circled{2}, to observe if the results obtained with the VSR were due to the lower number of parameters to optimize in that case, or if the \algnhl model is also able to scale to bigger NNs. 
In \Cref{fig:fitness-evolution-ant}, we show the fitness trend for the two networks we tested on the Ant task. 
In this case, the trends are slightly different: with the ``medium'' NN, the \algnhl model seems to converge faster, while with the ``high'' NN, the \algnhl model seems to converge slower than the \alghl model. 
However, we can observe that eventually, with both networks, the performance of \alghl and \algnhl converge to similar values. We confirmed this in \Cref{fig:rq1-bp-ant}, where we plot the average performance over $100$ task rollouts (as done in \cite{najarro2020meta}) of the best solutions of each independent run. 
The \algnhl models reach an average score of $1148 \pm 98$ and $1216 \pm 149$ for the medium and high configuration (in comparison, a standard RL method such as PPO reaches a score of around $3100$ \cite{RL2020ppo}). In this case, the TOST test returned a $p$-value of $0.26$ and $0.66$, respectively for the ``medium'' and ``high'' models. Hence also in this case we cannot reject the hypothesis of statistical equivalence.

\begin{figure}
 \centering
 \begin{tikzpicture}
 \begin{groupplot}[
 boxplot,
 boxplot/draw direction=y,
 width=1\linewidth,
 height=0.5\linewidth,
 group style={
 group size=1 by 1,
 horizontal sep=1mm,
 vertical sep=1.5mm,
 xticklabels at=edge bottom,
 yticklabels at=edge left,
 },
 ymin=-1,ymax=1200,
 legend cell align={left},
 ymajorgrids=true,
 yminorgrids=true,
 xtick={1.5, 3.5, 5.5, 7.5, 9.5, 11.5},
 xticklabels={{Worm\\Low}, {Worm\\Medium}, {Worm\\High}, {Biped\\Low}, {Biped\\Medium}, {Biped\\High}},
 ticklabel style={font=\footnotesize},
 grid style={line width=.1pt, draw=gray!10},
 major grid style={line width=.15pt, draw=gray!50},
 minor y tick num=4,
 title style={anchor=north, yshift=2ex},
 xmajorticks=true,
 xminorticks=true
 ]
 \nextgroupplot[
 align=center,
 legend columns=2,
 legend entries={\alghl, \algnhl},
 legend style={draw=none},
 ylabel={Distance},
 legend to name=legendRQ1box-vsr
 ]
 \addlegendimage{area legend,color=cola1,fill}
 \addlegendimage{area legend,color=cola2,fill}

 \addplot[black,fill=cola1] table[y=h] {data/rq1/bp/wl.txt};
 \addplot[black,fill=cola2] table[y=n] {data/rq1/bp/wl.txt};
 \pvalue{1}{2}{450}{50}{$0.39$}
 \addplot[black,fill=cola1] table[y=h] {data/rq1/bp/wm.txt};
 \addplot[black,fill=cola2] table[y=n] {data/rq1/bp/wm.txt};
 \pvalue{3}{4}{350}{50}{$0.07$}
 \addplot[black,fill=cola1] table[y=h] {data/rq1/bp/wh.txt};
 \addplot[black,fill=cola2] table[y=n] {data/rq1/bp/wh.txt};
 \pvalue{5}{6}{350}{50}{$0.20$}
 
 \addplot[black,fill=cola1] table[y=h] {data/rq1/bp/bl.txt};
 \addplot[black,fill=cola2] table[y=n] {data/rq1/bp/bl.txt};
 \pvalue{7}{8}{900}{50}{$0.79$}
 \addplot[black,fill=cola1] table[y=h] {data/rq1/bp/bm.txt};
 \addplot[black,fill=cola2] table[y=n] {data/rq1/bp/bm.txt};
 \pvalue{9}{10}{750}{50}{$0.30$}
 \addplot[black,fill=cola1] table[y=h] {data/rq1/bp/bh.txt};
 \addplot[black,fill=cola2] table[y=n] {data/rq1/bp/bh.txt};
 \pvalue{11}{12}{750}{50}{$0.96$}
 
 
 \draw [dashed] (6.5,\pgfkeysvalueof{/pgfplots/ymin}) -- (6.5,\pgfkeysvalueof{/pgfplots/ymax});
 
 \end{groupplot}
 \end{tikzpicture}
  \vspace*{-0.2cm}
 \caption{RQ1: Distribution of the best distance achieved across 10 runs by \alghl and \algnhl in the various configurations of the VSR task.}
 \Description{}
 \label{fig:rq1-bp-vsr}
 \vspace{-0.5cm}
\end{figure}
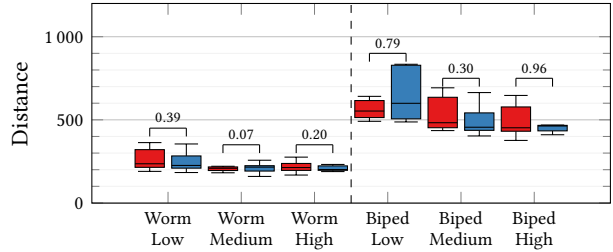

\begin{figure}[ht!]
 \centering
 \begin{tikzpicture}
 \begin{groupplot}[
 width=0.55\linewidth,
 height=0.45\linewidth,
 grid=both,
 grid style={line width=.1pt, draw=gray!10},
 major grid style={line width=.2pt,draw=gray!50},
 minor tick num=4,
 title style={anchor=north, yshift=2ex},
 group style={
 group size=2 by 1,
 horizontal sep=4mm,
 vertical sep=3mm,
 xticklabels at=edge bottom,
 yticklabels at=edge left
 },
 scaled x ticks = false,
 x tick label style={
 /pgf/number format/.cd,
 fixed,
 fixed zerofill,
 int detect,
 1000 sep={},
 precision=3
 },
 ylabel style={
 align=center
 },
 every axis plot/.append style={thick},
 ymin=0,ymax=1600,
 xmin=0,xmax=500,
 xtick={0,100,300,500}
 ]
 \nextgroupplot[
 legend entries={$1$ layer, $2$ layers, $3$ layers}, 
 legend to name=la,
 legend columns=3,
 legend style={draw=none},
 ylabel={Distance},
 xlabel={Generations},
 title={Medium}
 ]
 \addlegendimage{dotted,black}
 \addlegendimage{dashed,black}
 \addlegendimage{solid,black}
 \linewitherror{data/rq1/ft/amh.txt}{i}{m}{s}{cola1}
 \linewitherror{data/rq1/ft/amn.txt}{i}{m}{s}{cola2}
 
 \nextgroupplot[xlabel={Generations}, title={High}]
 
 \linewitherror{data/rq1/ft/abh.txt}{i}{m}{s}{cola1}
 \linewitherror{data/rq1/ft/abn.txt}{i}{m}{s}{cola2}
 \end{groupplot}
 \end{tikzpicture}
 \vspace*{-0.2cm}
 \caption{RQ1: Distance trend of the best solution (avg. $\pm$ std. dev. across 10 runs) found at each generation by \alghl and \algnhl in the two configurations of the Ant task.}
 \Description{}
 \label{fig:fitness-evolution-ant}
 \vspace{-0.4cm}
\end{figure}
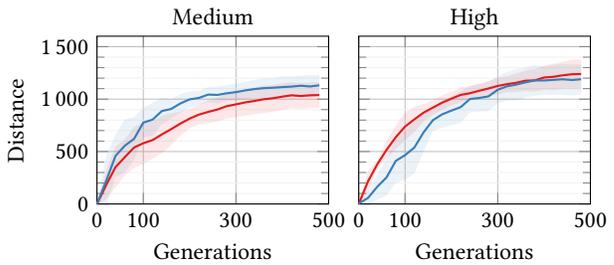

\subsubsection*{\textbf{RQ2: \algwnhl performance}}

Aiming to answer the second research question (RQ2), we focused only on the hardest task, i.e., Ant. 
The only difference w.r.t. to the previous experimental setting is that, in this case, the weights of \algnhl are initialized to $0$ at the beginning of the task, for a fair comparison with \algwnhl where, instead, as seen earlier we assume by construction that there are no initial weights, which allows us to avoid storing them.

For the optimization process, in this set of experiments, we opted for ES$_1$, with the same parametrization as the previous experiments, preferred over ES$_2$ due to computational constraints (in fact, it uses a population size that is $12.5$ times smaller). 

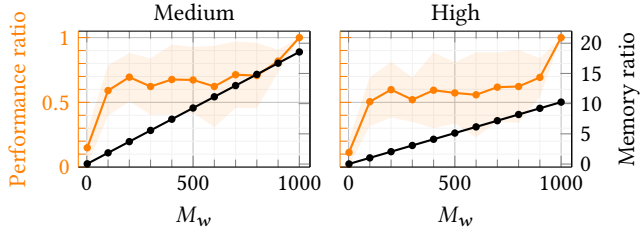
\begin{figure}[ht!]
\vspace{-0.25cm}
 \centering
 \begin{tikzpicture}
 \begin{groupplot}[
 at={(1.25cm,-3.5cm)},
 width=0.55\linewidth,
 height=0.4\linewidth,
 grid=both,
 grid style={line width=.1pt, draw=gray!10},
 major grid style={line width=.2pt, draw=gray!50},
 minor tick num=4,
 title style={anchor=north, yshift=2ex},
 group style={
 group size=2 by 1,
 horizontal sep=4mm,
 vertical sep=3mm,
 xticklabels at=edge bottom,
 yticklabels at=edge left
 },
 scaled x ticks = false,
 ytick pos=left,
 x tick label style={
 /pgf/number format/.cd,
 fixed,
 fixed zerofill,
 int detect,
 1000 sep={},
 precision=3,
 ylabel style={cola5},
 ytick style={cola5},
 every y tick label/.append style={cola5},
 separate axis lines,
 first y axis line style={cola5}
 },
 ymin=0,ymax=1.05,
 xlabel={$M_w$},
 xmin=-40,xmax=1050
 ]
 \nextgroupplot[
 ylabel style={cola5},
 ylabel={Performance ratio},
 title={Medium}
 ]{
 \linewitherrorns{data/rq2/val/val_med.txt}{i}{m}{s}{cola5}
 }
 \nextgroupplot[
 xlabel={$M_w$},
 title={High}
 ]{
 \linewitherrorns{data/rq2/val/val_big.txt}{i}{m}{s}{cola5}
 }
 \end{groupplot}
 \begin{groupplot}[
 at={(1.25cm,-3.5cm)},
 width=0.55\linewidth,
 height=0.4\linewidth,
 group style={
 group size=2 by 1,
 horizontal sep=4mm,
 vertical sep=3mm,
 xticklabels at=edge bottom,
 yticklabels at=edge left
 },
 axis y line*=right,
 axis x line=none,
 ymin=-0.5,ymax=22,
 xmin=-40,xmax=1050,
 legend pos=south east
 ]
 \nextgroupplot[ytick={0,5,10,15,20,22},
 yticklabels={}]
 {
 \linesimple{data/rq2/val/val_med.txt}{i}{r}{black}
 }
 \nextgroupplot[
 xlabel={$M_w$},
 ylabel={Memory ratio},
 ytick={0,5,10,15,20,22},
 yticklabels={0,5,10,15,20}]{
 \linesimple{data/rq2/val/val_big.txt}{i}{r}{black}
 }
 \end{groupplot}
 \end{tikzpicture}
 \vspace{-0.7cm}
 \caption{RQ2: Ratio of the performance of the \algnhl model w.r.t. the \algwnhl model (orange line) with different memory windows $M_w$ (avg. $\pm$ std. dev. across the best solution found in each of $10$ runs of \algnhl, and then converted it to the corresponding \algwnhl model, on the Ant task). We also indicate the ratio between the memory consumption in terms of the number of activations (for \algwnhl) and the number of weights (for \algnhl) in two models (black line).}
 \Description{}
 \label{fig:approxTrend}
 \vspace{-0.3cm}
\end{figure}

In a first set of experiments, we evolved the \algnhl models for $10$ independent runs. We then took the best solution found in each run, and, given that the initial weights are set to $0$, we applied the \algwnhl model, see \Cref{eq:weightless}, to each of them. We did that for various sizes of the memory window ($M_w$) on which the model approximates the weights, to assess its effect on performance. Taking as the biggest possible window the maximum duration of the task (which, in the case of Ant, is $1000$ forward steps), we considered progressively increasing values of $M_w$, from $2$ to $1000$, with equal spacing. 
We show the average results obtained in \Cref{fig:approxTrend}. In the figure, the orange line shows the ratio between the fitness of the \algwnhl model and the fitness of the corresponding \algnhl model, namely $F_{ratio} = \frac{F_{\algwnhl}}{F_{\algnhl}}$, while varying $M_w$. 
The black line, instead, shows the ratio between the number of activations stored in \algwnhl and the number of weights stored in \algnhl. 

We can observe that, as expected, increasing the memory window increases the performance of the \algwnhl model, eventually reaching the same performance as the \algnhl model ($F_{ratio}=1$) when storing all the activations ($M_w=1000$). However, the memory needed to store the activations values, on which the weights are approximated in \algwnhl, rapidly becomes much bigger than the memory needed to directly store the weights, thus reducing the memory advantage of the \algwnhl approach over the \algnhl model.

\begin{figure}[ht!]
\vspace{-0.25cm}
 \centering
 \begin{tikzpicture}
 \begin{groupplot}[
 width=0.55\linewidth,
 height=0.4\linewidth,
 grid=both,
 grid style={line width=.1pt, draw=gray!10},
 major grid style={line width=.2pt,draw=gray!50},
 minor tick num=4,
 title style={anchor=north, yshift=2ex},
 group style={
 group size=2 by 1,
 horizontal sep=4mm,
 vertical sep=3mm,
 xticklabels at=edge bottom,
 yticklabels at=edge left
 },
 scaled x ticks = false,
 x tick label style={
 /pgf/number format/.cd,
 fixed,
 fixed zerofill,
 int detect,
 1000 sep={},
 precision=3
 },
 ylabel style={
 align=center
 },
 every axis plot/.append style={thick},
 ymin=0,ymax=1700,
 xmin=0,xmax=500,
 ]
 \nextgroupplot[
 title={Medium},
 ylabel={Distance},
 xlabel={Generation}
 ]
 \linewitherror{data/rq2/medium/nnn.txt}{i}{m}{s}{cola2}
 \linewitherror{data/rq2/medium/whl.txt}{i}{m}{s}{cola3}
 \nextgroupplot[
 title={High},
 ylabel={},
 xlabel={Generation}
 ]
 \linewitherror{data/rq2/big/nnn.txt}{i}{m}{s}{cola2}
 \linewitherror{data/rq2/big/whl.txt}{i}{m}{s}{cola3}
 \end{groupplot}
 \end{tikzpicture}
 \vspace{-0.2cm}
 \caption{RQ2: Distance trend of the best solution (avg. $\pm$ std. dev. across 10 runs) found at each generation by \algnhl and \algwnhl in the two configurations of the Ant task.}
 \Description{}
 \label{fig:rq2_fit_trend}
 \vspace{-0.4cm}
\end{figure}
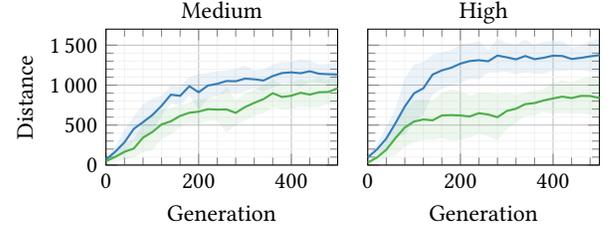

Therefore, given the relevant drop in memory efficiency for large values of the memory window, we performed a second experiment, where we set $M_w = 2$, and optimized the Hebbian parameters for this \algwnhl model. 
We show these results in \Cref{fig:rq2_fit_trend}. While the performance of the \algwnhl model does not reach the same performance of the \algnhl model (we verified this with the Mann-Whitney U test, which returned a $p$-value equal to $0.009$ and $0.008$, respectively for the ``medium'' and ``high'' configuration), we can see how the \algwnhl model still reaches fairly good performances, especially in the ``medium'' case. Moreover, optimizing the Hebbian parameters leads to an increase in the $F_{ratio}$ from $0.14$ and $0.11$ to $0.83$ and $0.64$, respectively for the ``medium'' and ``high'' models. 


Summarizing, we observed that, using, for the ``medium'' network, only $M_{ratio}=\frac{M_{w}\times N}{W}=\frac{2\times228}{12880}=\sim0.03\%$ of the memory of \algnhl, we lost only $\sim17\%$ of performance (see the final distance achieved by the two models shown in \Cref{fig:rq2_fit_trend}). Correspondingly, for the ``high'' network the memory consumption of \algwnhl is $\sim0.02\%$ of \algnhl with a performance drop of $\sim37\%$. While this performance drop can be considered significant, there might be cases where this performance-memory trade-off could be acceptable.
For instance, the ``high'' network with the \algnhl model needs to store $40960$ weights, $2100$ Hebbian rule parameters, and $392$ activations, for a total of $43452$ values. Considering $32$-bit float-point representation, this model would need $\sim170$kB (as a concrete example, an Arduino Nano has only $30$kB of Flash Memory available). In comparison, the \algwnhl model with $M_w=2$ requires storing $2$ activations and the $5$ Hebbian rule parameters for each neuron (for a total $420$ neurons), hence requiring $\sim11$kB of memory. Therefore, \algnhl requires about $15$ times more memory than \algwnhl.


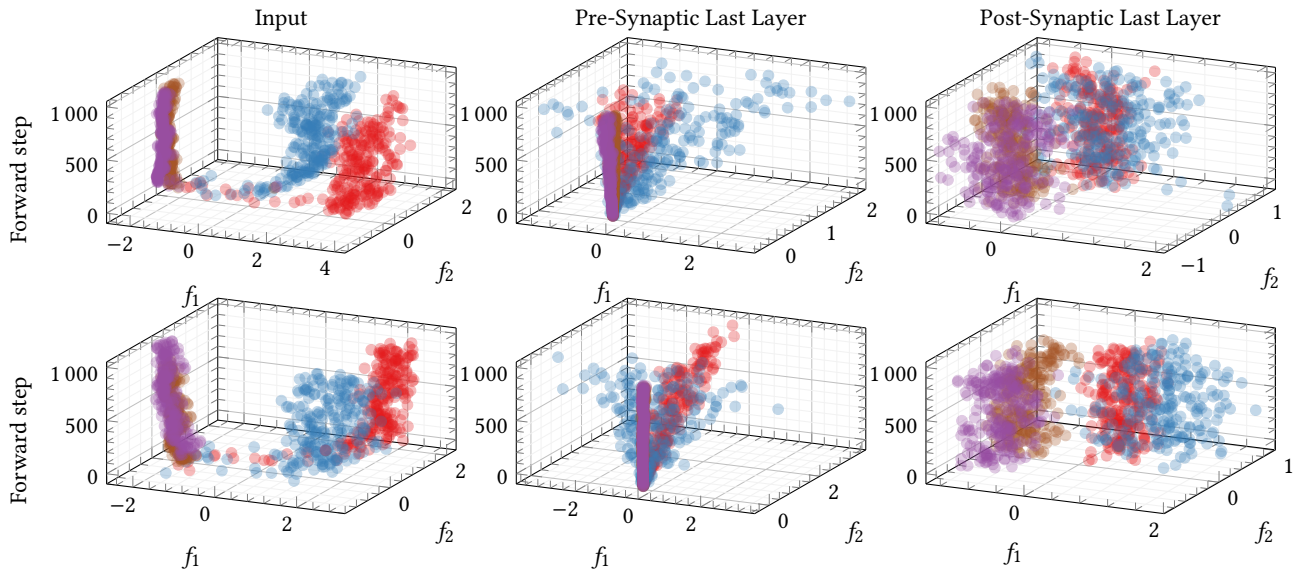
\begin{figure*}[ht!]
\centering
\begin{tikzpicture}
 \begin{groupplot}[
 width=0.35\linewidth,
 height=0.25\linewidth,
 grid=both,
 grid style={line width=.1pt, draw=gray!10},
 major grid style={line width=.2pt,draw=gray!50},
 minor tick num=4,
 title style={anchor=north, yshift=2ex},
 group style={
 group size=3 by 2,
 horizontal sep=8mm,
 vertical sep=6mm
 },
 ylabel style={
 align=center
 },
 zlabel style={
 align=center
 },
 xlabel style={
 align=center
 }
 ]
 \nextgroupplot[title={Input}, xlabel={$f_1$}, ylabel={$f_2$}, zlabel={Forward step}]{
 \addplot3[
 scatter/classes={
 neuron={mark=*, cola2},
 hebbian={mark=*, cola1},
 hebbianRandom={mark=*, cola7},
 neuronRandom={mark=*, cola4},
 sth={mark=+, cola2}
 },
 scatter,
 only marks,
 scatter src=explicit symbolic,
 opacity=0.3,
 each nth point={4}
 ]
 table[x=x,y=y,z=i,meta=label] {data/rq3/mean3/binp.txt};
 } 
 \nextgroupplot[title={Pre-Synaptic Last Layer}, xlabel={$f_1$}, ylabel={$f_2$}]{
 \addplot3[
 scatter/classes={
 neuron={mark=*, cola2},
 hebbian={mark=*, cola1},
 hebbianRandom={mark=*, cola7},
 neuronRandom={mark=*, cola4}
 },
 scatter,
 only marks,
 scatter src=explicit symbolic,
 opacity=0.3,
 each nth point={4}
 ]
 table[x=x,y=y,z=i,meta=label] {data/rq3/mean3/bpre.txt};
 } 
 \nextgroupplot[title={Post-Synaptic Last Layer}, xlabel={$f_1$}, ylabel={$f_2$}]{
 \addplot3[
 scatter/classes={
 neuron={mark=*, cola2},
 hebbian={mark=*, cola1},
 hebbianRandom={mark=*, cola7},
 neuronRandom={mark=*, cola4}
 },
 scatter,
 only marks,
 scatter src=explicit symbolic,
 opacity=0.3,
 each nth point={4}
 ]
 table[x=x,y=y,z=i,meta=label] {data/rq3/mean3/bout.txt};
 } 
 \nextgroupplot[xlabel={$f_1$}, ylabel={$f_2$}, zlabel={Forward step}]{
 \addplot3[
 scatter/classes={
 neuron={mark=*, cola2},
 hebbian={mark=*, cola1},
 hebbianRandom={mark=*, cola7},
 neuronRandom={mark=*, cola4}
 },
 scatter,
 only marks,
 scatter src=explicit symbolic,
 opacity=0.3,
 each nth point={4},
 ]
 table[x=x,y=y,z=i,meta=label] {data/rq3/mean3/minp.txt};
 } 
 \nextgroupplot[xlabel={$f_1$}, ylabel={$f_2$}]{
 \addplot3[
 scatter/classes={
 neuron={mark=*, cola2},
 hebbian={mark=*, cola1},
 hebbianRandom={mark=*, cola7},
 neuronRandom={mark=*, cola4}
 },
 scatter,
 only marks,
 scatter src=explicit symbolic,
 opacity=0.3,
 each nth point={4},
 ]
 table[x=x,y=y,z=i,meta=label] {data/rq3/mean3/mpre.txt};
 } 
 \nextgroupplot[xlabel={$f_1$}, ylabel={$f_2$}, 
 each nth point={4},
 legend columns=4,
 legend entries={\alghl, \algnhl, \alghl-Random, \algnhl-Random},
 legend style={draw=none, mark options={opacity=1.}},
 legend to name=legendPca,
 ]{
 \addplot3[
 scatter/classes={
 neuron={mark=*, cola2},
 hebbian={mark=*, cola1},
 hebbianRandom={mark=*, cola7},
 neuronRandom={mark=*, cola4}
 },
 scatter,
 only marks,
 scatter src=explicit symbolic,
 opacity=0.3
 ]
 table[x=x,y=y,z=i,meta=label] {data/rq3/mean3/mout.txt};
 
 \addlegendimage{area legend,color=cola1,fill}
 \addlegendimage{area legend,color=cola2,fill}
 \addlegendimage{area legend,color=cola7,fill}
 \addlegendimage{area legend,color=cola4,fill}
 }
 \end{groupplot}
 \end{tikzpicture}
 \vspace{-0.3cm}
 \caption{RQ3: Average trajectory at various network layers (Input layer; Pre-Synaptic Last Layer; Post-Synaptic Last Layer) of the best solutions found on the Ant task (top row: Medium; bottom row: High) from 3 rollouts of each of the best solutions found in 10 evolutionary runs of \alghl, \algnhl, \textcolor{cola7}{HL-Random}, and \textcolor{cola4}{NcHL-Random}, each one tested in 3 rollouts. Each point is a 2D projection (through PCA) of an n-dimensional vector 
 corresponding to each of the shown layers at a certain forward step.} 
 \Description{}
 \label{fig:pca}
 \vspace{-0.4cm}
\end{figure*}

\subsubsection*{\textbf{RQ3: Behavior analysis}}

To conclude our experiments, we performed a qualitative behavioral analysis of the final networks obtained with both \alghl and \algnhl. As shown in our RQ1 experiments, the final fitness results, as well as their overall trend, for the networks trained with these two different approaches turn out to be very close. Hence, we further investigated the reason behind such a trend. For this analysis, we considered only the most difficult task, i.e., Ant, because the bigger-sized networks used to solve it (see the ``medium'' and ``high'' NNs in \Cref{tab:task_p}) should be able to provide more information on the models' behavior.

The first step was to analyze the similarity among the activations of \alghl and \algnhl. However, we discovered that, after just a few steps, all the activations in both models go to the maximum/minimum values of $tanh$ (i.e., $-1$ or $1$). We hypothesized that such behavior on the activations is due to the lack of normalization of the network weights (which was done following the setting of the original study from which the Ant task is taken \cite{najarro2020meta}). 
Of note, this pattern can however be linked to several biological studies \cite{eckert1988animal,kalat2016biological} that discuss the ``all-or-nothing'' behavior of biological neurons, i.e., the fact that such neurons do not have an intermediate activation value: they either fire, or not.

Hence, we decided to better analyze the network similarity by relying on the behavioral trajectories.
For each of the best solutions found in each of $10$ evolutionary runs on \alghl and \algnhl for RQ1, we extracted the trajectories from $3$ test rollouts. Likewise, we created $10$ individuals for both \alghl-Random and \algnhl-Random, and tested them over $3$ rollouts.
For each trajectory, we stored three pieces of information for every forward step (the total number of steps being $1000$), namely the input values $I \in \mathbb{R}^{1\times28}$, the pre-synaptic $P \in \mathbb{R}^{1\times8}$, and the post-synaptic activations of the last layer $P' \in \mathbb{R}^{1\times8}$.
Note that with pre-synaptic activations, we mean the values of the output layer before the activation function, hence the values are in the range $[-\infty, \infty]$. While with post-synaptic activations, we mean the values after the activation function, hence they are in the range $[-1,1]$.


We conducted this analysis performing, for each of the $30$ trajectories of each model, a PCA for each step, separately over the input, pre-synaptic, and post-synaptic values. Then, we calculated for each step the average over all the PCA results of the $30$ trajectories.

The averaged trajectories of each algorithm, shown in \Cref{fig:pca} indicate that: (1) \alghl and \algnhl explore the behavioral space more than the random baselines; (2) the \alghl and \algnhl trajectories turn out to overlap, being separate from the random cases. Analyzing such results from left (input) to right (post-synaptic) over \Cref{fig:pca} (top row: ``medium'' network; bottom row: ``high'' network), it is clear how the input trajectories of \alghl and \algnhl are similar in the ``medium'' networks, while they almost completely overlap in the ``high'' case. On the contrary, the random baselines overlap in both cases and remain close to the starting 2D point (the z-axis corresponds to the forward steps). A similar trend is visible in the second column, where the random baselines move along a straight line over the forward axis, while \alghl and \algnhl follow a more complex, but similar trajectory. In the last column, the post-synaptic values (which, as discussed before, turn out to be either $-1$ or $1$), confirm that \alghl and \algnhl explore the behavioral space similarly, and more than the random baselines. 

These findings allow us to empirically show one of the reasons behind the similar performance obtained by \alghl and \algnhl over the experiments reported in RQ1. Although our proposed \algnhl significantly reduces the number of parameters to be optimized, its ability to explore the behavior space turns out to be as effective as that of the traditional, but more parametrized ABCD rule.
\vspace{-0.3cm}


\section{Conclusions}
\label{sec:conclusions}

In this paper, we have introduced a novel variant of \alghl that focuses on neurons rather than synapses. We called this model \algdesc (\algnhl), to highlight the fact that the Hebbian update rule is focused, by design, on neurons. 
We then proposed a further ``weightless'' model (\algwnhl), that allows for plasticity but without the need for storing weights. We tested the proposed models on two simulated robotic locomotion tasks in various configurations, showing their effectiveness both in terms of performance and number of parameters. Our experiments showed that, despite a massive reduction in the overall number of parameters to optimize w.r.t. the traditional \alghl model, \algnhl provides comparable performance on the two tasks at hand. On the other hand, \algwnhl resulted slightly less competitive than \algnhl, which reveals its trade-off between performance and memory-saving.

We concluded our study with a behavioral characterization of \alghl and \algnhl, in comparison with two random baselines. This analysis provided some interesting insights as to what concerns the way \algnhl explores the behavioral space following somehow similar trajectories w.r.t. \alghl, but differently from the random baselines.

In our opinion, this work could represent a first step towards a radical rethinking of \alghl, since as said we put the emphasis of the Hebbian update on the neurons rather than synapses. To the best of our knowledge, no prior work proposed such a way of formulating \alghl. Furthermore, our work allows us to preserve a biologically plausible model while reducing the number of parameters, thus paving the way to scalable \alghl models. 
Of note, in this work we did not compare our method with alternative approaches to model size reduction, or more advanced HL models: given that our goal was to find a novel way to achieve scalability (from the optimization viewpoint) in HL, rather than improve the performance of HL, we thought that the most meaningful baseline was the vanilla Synaptic-centric HL. Therefore, further investigations are needed to assess the performance of \algnhl, also in comparison with alternative learning approaches, and its generalizability e.g. to deep learning scenarios and complex high-dimensional tasks such as computer vision. 


\clearpage

\balance

\bibliographystyle{ACM-Reference-Format}
\bibliography{main}

\end{document}

%% file: drawing.tex
\usetikzlibrary{positioning}

\pgfplotsset{
    every first x axis line/.style={},
    every first y axis line/.style={},
    every first z axis line/.style={},
    every second x axis line/.style={},
    every second y axis line/.style={},
    every second z axis line/.style={},
    first x axis line style/.style={/pgfplots/every first x axis line/.append style={#1}},
    first y axis line style/.style={/pgfplots/every first y axis line/.append style={#1}},
    first z axis line style/.style={/pgfplots/every first z axis line/.append style={#1}},
    second x axis line style/.style={/pgfplots/every second x axis line/.append style={#1}},
    second y axis line style/.style={/pgfplots/every second y axis line/.append style={#1}},
    second z axis line style/.style={/pgfplots/every second z axis line/.append style={#1}}
}

\makeatletter
\def\pgfplots@drawaxis@outerlines@separate@onorientedsurf#1#2{%
    \if2\csname pgfplots@#1axislinesnum\endcsname
    \else
    \scope[/pgfplots/every outer #1 axis line,
        #1discont,decoration={pre length=\csname #1disstart\endcsname, post length=\csname #1disend\endcsname}]
        \pgfplots@ifaxisline@B@onorientedsurf@should@be@drawn{0}{%
            \draw [/pgfplots/every first #1 axis line] decorate {
                \pgfextra
                \pgfplotspointonorientedsurfaceabsetupfor{#2}{#1}{\pgfplotspointonorientedsurfaceN}%
                \pgfplots@drawgridlines@onorientedsurf@fromto{\csname pgfplots@#2min\endcsname}%
                \endpgfextra 
                };
        }{}%
        \pgfplots@ifaxisline@B@onorientedsurf@should@be@drawn{1}{%
            \draw [/pgfplots/every second #1 axis line] decorate {
                \pgfextra
                \pgfplotspointonorientedsurfaceabsetupfor{#2}{#1}{\pgfplotspointonorientedsurfaceN}%
                \pgfplots@drawgridlines@onorientedsurf@fromto{\csname pgfplots@#2max\endcsname}%
                \endpgfextra 
                };
        }{}%
    \endscope
    \fi
}%
\makeatother

\newcommand{\vt}[1]{\adjustbox{}{\tikz{#1}}}

\newcommand{\sensors}[9]{ 
    \ifthenelse{#3 = 100}{
        \filldraw[fill=red!#3, draw=black!#3] (\l/5+#1*\l -\l/15 , \l/2+#2*\l - \l/15) rectangle (\l/5+#1*\l +\l/15 , \l/2+#2*\l + \l/15);
    }{
    \pgfmathsetmacro{\cnt}{#3+ 50}
        \filldraw[fill=black!0, draw=black!\cnt] (\l/5+#1*\l -\l/15 , \l/2+#2*\l - \l/15) rectangle (\l/5+#1*\l +\l/15 , \l/2+#2*\l + \l/15);
    }
    \ifthenelse{#4 = 100}{
        \filldraw[fill=red!#4, draw=black!#4]  (2*\l/5+#1*\l-\l/15, \l/2+#2*\l-\l/15) rectangle (2*\l/5+#1*\l+\l/15, \l/2+#2*\l+\l/15);
    }{
    \pgfmathsetmacro{\cnt}{#4+ 50}
        \filldraw[fill=black!0, draw = black!\cnt] (2*\l/5+#1*\l-\l/15, \l/2+#2*\l-\l/15) rectangle (2*\l/5+#1*\l+\l/15, \l/2+#2*\l+\l/15);
    }
    \ifthenelse{#5 = 100}{
        \filldraw[fill=red!#5, draw=black!#5] (3*\l/5+#1*\l-\l/15, \l/2+#2*\l) rectangle (3*\l/5+#1*\l+\l/15, \l/2+#2*\l+2*\l/15);
    }{
    \pgfmathsetmacro{\cnt}{#5+ 50}
        \filldraw[fill=black!0, draw=black!\cnt] (3*\l/5+#1*\l-\l/15, \l/2+#2*\l) rectangle (3*\l/5+#1*\l+\l/15, \l/2+#2*\l+2*\l/15);
    }
    \ifthenelse{#6 = 100}{
        \filldraw[fill=red!#6, draw=black!#6] (3*\l/5+#1*\l-\l/15, \l/2+#2*\l) rectangle (3*\l/5+#1*\l+\l/15, \l/2+#2*\l-2*\l/15);
    }{
    \pgfmathsetmacro{\cnt}{#6+ 50}
       \filldraw[fill=black!0, draw=black!\cnt] (3*\l/5+#1*\l-\l/15, \l/2+#2*\l) rectangle (3*\l/5+#1*\l+\l/15, \l/2+#2*\l-2*\l/15);
    }
    \ifthenelse{#7 = 100}{
        \filldraw[fill=red!#7, draw=black!#7] (4*\l/5+#1*\l-\l/15, \l/2+#2*\l+\l/15) rectangle (4*\l/5+#1*\l+\l/15, \l/2+#2*\l+\l/5); 
    }{
    \pgfmathsetmacro{\cnt}{#7+ 50}
        \filldraw[fill=black!0, draw=black!\cnt] (4*\l/5+#1*\l-\l/15, \l/2+#2*\l+\l/15) rectangle (4*\l/5+#1*\l+\l/15, \l/2+#2*\l+\l/5); 
    }
    \ifthenelse{#8 = 100}{
        \filldraw[fill=red!#8, draw=black!#8] (4*\l/5+#1*\l-\l/15, \l/2+#2*\l-\l/15) rectangle (4*\l/5+#1*\l+\l/15, \l/2+#2*\l+\l/15);
    }{
    \pgfmathsetmacro{\cnt}{#8+ 50}
        \filldraw[fill=black!0, draw=black!\cnt] (4*\l/5+#1*\l-\l/15, \l/2+#2*\l-\l/15) rectangle (4*\l/5+#1*\l+\l/15, \l/2+#2*\l+\l/15);
    }
    \ifthenelse{#9 = 100}{
        \filldraw[fill=red!#9, draw=black!#9] (4*\l/5+#1*\l-\l/15, \l/2+#2*\l-\l/15) rectangle (4*\l/5+#1*\l+\l/15, \l/2+#2*\l-\l/5);
    }{
        \pgfmathsetmacro{\cnt}{#9 + 50}
        \filldraw[fill=black!0, draw=black!\cnt] (4*\l/5+#1*\l-\l/15, \l/2+#2*\l-\l/15) rectangle (4*\l/5+#1*\l+\l/15, \l/2+#2*\l-\l/5);
    }
}

\newcommand{\ndrcasec}[2]{
    \node[circle, draw] (I) at (0*#1,1*#2){I}; 
    
    \node[circle, draw] (A) at (1*#1,0*#2){A};
    \node[circle, draw] (B) at (1*#1,1*#2){B};
    \node[circle, draw] (C) at (1*#1,2*#2){C};
    \node[circle, draw] (O) at (2*#1,1*#2){O};
    
    \node[circle] (D) at (1*#1,3*#2){};
    
    \draw[->] (I) to[right] node[left]{}(B) ;
    \draw[->] (I) to[right] node[left]{}(A) ;
    \draw[->, red] (I) to[right] node[left]{}(C) ;
    
    \draw[->, red] (A) to[right] node[left]{}(O) ;
    \draw[->, red] (B) to[right] node[left]{}(O) ;
    \draw[->] (C) to[right] node[left]{}(O) ;
    
    \draw[<->, red] (A) to[right] node[left]{}(B) ;
    \draw[->] (B) to[right] node[left]{}(C) ;
    \draw[->] (A) to[bend right] node[left]{}(C) ;
    
    \draw[-] (I) to[bend left=60] node[left]{}(D.west);
    \draw[-] (D.east) to node[left]{}(D.west);
    \draw[->] (D.east) to[bend left=60] node[left]{}(O);
}
\tikzset{pics/ndrcasecpc/.style n args={2}{code={
    \ndrcasec{#1}{#2}
}}}

\newcommand{\ndrcaseb}[2]{
    \node[circle, draw] (I) at (0*#1,1*#2){I}; 
    
    \node[circle, draw] (A) at (1*#1,0*#2){A};
    \node[circle, draw] (B) at (1*#1,1*#2){B};
    \node[circle, draw] (C) at (1*#1,2*#2){C};
    \node[circle, draw] (O) at (2*#1,1*#2){O};
    
    \node[circle] (D) at (1*#1,3*#2){};
    
    \draw[->, red] (I) to[right] node[left]{}(B) ;
    \draw[->, red] (I) to[right] node[left]{}(A) ;
    \draw[->, red] (I) to[right] node[left]{}(C) ;
    
    \draw[->, red] (A) to[right] node[left]{}(O) ;
    \draw[->, red] (B) to[right] node[left]{}(O) ;
    \draw[->, red] (C) to[right] node[left]{}(O) ;
    
    \draw[<->, red] (A) to[right] node[left]{}(B) ;
    \draw[<->, red] (B) to[right] node[left]{}(C) ;
    \draw[<->, red] (A) to[bend right] node[left]{}(C) ;
    
    \draw[-] (I) to[bend left=60] node[left]{}(D.west);
    \draw[-] (D.east) to node[left]{}(D.west);
    \draw[->] (D.east) to[bend left=60] node[left]{}(O);
}
\tikzset{pics/ndrcasebpc/.style n args={2}{code={
    \ndrcaseb{#1}{#2}
}}}

\newcommand{\ndrcasea}[2]{
    \node[circle, draw] (I) at (0*#1,1*#2){I}; 
    
    \node[circle, draw] (A) at (1*#1,0*#2){A};
    \node[circle, draw] (B) at (1*#1,1*#2){B};
    \node[circle, draw] (C) at (1*#1,2*#2){C};
    \node[circle, draw] (O) at (2*#1,1*#2){O};
    
    \node[circle] (D) at (1*#1,3*#2){};
    
    \draw[->] (I) to[right] node[left]{}(B) ;
    \draw[->] (I) to[right] node[left]{}(A) ;
    \draw[->] (I) to[right] node[left]{}(C) ;
    
    \draw[->] (A) to[right] node[left]{}(O) ;
    \draw[->] (B) to[right] node[left]{}(O) ;
    \draw[->] (C) to[right] node[left]{}(O) ;
    
    \draw[<->, red] (A) to[right] node[left]{}(B) ;
    \draw[<->, red] (B) to[right] node[left]{}(C) ;
    \draw[<->, red] (A) to[bend right] node[left]{}(C) ;
    
    \draw[-] (I) to[bend left=60] node[left]{}(D.west);
    \draw[-] (D.east) to node[left]{}(D.west);
    \draw[->] (D.east) to[bend left=60] node[left]{}(O);
}

\tikzset{pics/ndrcaseapc/.style n args={2}{code={
    \ndrcasea{#1}{#2}
}}}

\tikzset{pics/ndrStart/.style n args={2}{code={
        \node[circle, draw] (I) at (0*#1,1*#2){I}; 
        \node[circle, draw] (A) at (1*#1,0){A};
        \node[circle, draw] (B) at (1*#1,1*#2){B};
        \node[circle, draw] (C) at (1*#1,2*#2){C};
        \node[circle, draw] (O) at (2*#1,1*#2){O};
        
        \node[circle] (D) at (1*#1,3*#2){};
        
        \draw[->, red] (I) to[right] node[left]{}(B) ;
        \draw[->, red] (I) to[right] node[left]{}(A) ;
        \draw[->, red] (I) to[right] node[left]{}(C) ;
        
        \draw[->, red] (A) to[right] node[left]{}(O) ;
        \draw[->, red] (B) to[right] node[left]{}(O) ;
        \draw[->, red] (C) to[right] node[left]{}(O) ;
    
        \draw[<->, red] (A) to[right] node[left]{}(B) ;
        \draw[<->, red] (B) to[right] node[left]{}(C) ;%
        \draw[<->, red] (A) to[bend right] node[left]{}(C) ;
        
        \draw[-, red] (I) to[bend left=60] node[left]{}(D.west);
        \draw[-, red] (D.east) to node[left]{}(D.west);
        \draw[->, red] (D.east) to[bend left=60] node[left]{}(O);
}}}

\tikzset{pics/ndr1step/.style n args={3}{code={
        \node[circle, draw] (I) at (0*#1,1*#2){I}; 
        \node[circle, draw] (A) at (1*#1,0){A};
        \node[circle, draw] (B) at (1*#1,1*#2){B};
        \node[circle, draw] (C) at (1*#1,2*#2){C};
        \node[circle, draw] (O) at (2*#1,1*#2){O};
        
        \node[circle] (D) at (1*#1,3*#2){};
        
        \draw[->, line width=0.5mm*#3] (I) to[right] node[left]{}(B) ;
        \draw[->, line width=0.12mm*#3] (I) to[right] node[left]{}(A) ;
        \draw[->, line width=0.15mm*#3] (I) to[right] node[left]{}(C) ;
        
        \draw[->, line width=0.6mm*#3] (A) to[right] node[left]{}(O) ;
        \draw[->, line width=0.02mm*#3] (B) to[right] node[left]{}(O) ;
        \draw[->, line width=0.6mm*#3] (C) to[right] node[left]{}(O) ;
    
        \draw[<->, line width=0.35mm*#3] (A) to[right] node[left]{}(B) ;
        \draw[<->, line width=0.43mm*#3] (B) to[right] node[left]{}(C) ;%
        \draw[<->, line width=0.23mm*#3] (A) to[bend right] node[left]{}(C) ;
        
        \draw[-, line width=0.03mm*#3] (I) to[bend left=60] node[left]{}(D.west);
        \draw[-, line width=0.03mm*#3] (D.east) to node[left]{}(D.west);
        \draw[->, line width=0.03mm*#3] (D.east) to[bend left=60] node[left]{}(O);
}}}
        
\tikzset{pics/ndrpp/.style n args={3}{code={
        \node[circle, draw] (I) at (0*#1,1*#2){I}; 
        \node[circle, draw] (A) at (1*#1,0){A};
        \node[circle, draw] (B) at (1*#1,1*#2){B};
        \node[circle, draw] (C) at (1*#1,2*#2){C};
        \node[circle, draw] (O) at (2*#1,1*#2){O};
        
        \node[circle] (D) at (1*#1,3*#2){};
        
        \draw[->, line width=0.5mm*#3] (I) to[right] node[left]{}(B) ;
        \draw[->, red] (I) to[right] node[left]{}(A) ;
        \draw[->, red] (I) to[right] node[left]{}(C) ;
        
        \draw[->, line width=0.6mm*#3] (A) to[right] node[left]{}(O) ;
        \draw[->, red] (B) to[right] node[left]{}(O) ;
        \draw[->, line width=0.6mm*#3] (C) to[right] node[left]{}(O) ;
    
        \draw[<->, line width=0.35mm*#3] (A) to[right] node[left]{}(B) ;
        \draw[<->, line width=0.43mm*#3] (B) to[right] node[left]{}(C) ;%
        \draw[<->, line width=0.23mm*#3] (A) to[bend right] node[left]{}(C) ;
        
        \draw[-, red] (I) to[bend left=60] node[left]{}(D.west);
        \draw[-, red] (D.east) to node[left]{}(D.west);
        \draw[->, red] (D.east) to[bend left=60] node[left]{}(O);
}}}

\tikzset{pics/hier/.style n args={1}{code={
            \def\l{#1} 
            \foreach \x in {0,1,2,3,4,5}
                \draw (0+\l,0+\x) rectangle (\l+\l,\l+\x) node[pos=.5] {$i_{\x}$};
            \draw [decorate, decoration = {calligraphic brace, mirror,amplitude=10pt}] (2.5,4.1) --  (2.5,5.9);
            \draw [decorate, decoration = {calligraphic brace, mirror,amplitude=10pt}] (2.5,2.1) --  (2.5,3.9);
            \draw [decorate, decoration = {calligraphic brace, mirror,amplitude=10pt}] (2.5,0.1) --  (2.5,1.9);
            
            \node[rectangle, draw] (a) at (3.5, 1){$SBM_{0}$};
            \node[rectangle, draw] (b) at (3.5, 3){$SBM_{1}$};
            \node[rectangle, draw] (c) at (3.5, 5){$SBM_{2}$};

            \node[rectangle, draw,align=center] (head) at (6.5, 3) {$Interpretable$ \\ $Model$};
            \node[rectangle, draw] (leaf2) at (9, 3) {$output$};
            
            \draw[->, shorten <=2pt,shorten >=2pt,>=stealth] (a.east) to[right] node[left]{}(head.west) ;
            \draw[->, shorten <=2pt,shorten >=2pt,>=stealth] (b.east) to[right] node[left]{}(head.west) ;
            \draw[->, shorten <=2pt,shorten >=2pt,>=stealth] (c.east) to[right] node[left]{}(head.west) ;
            
            \draw[->, shorten <=2pt,shorten >=2pt,>=stealth] (head) to[right] node[left]{}(leaf2.west) ;
}}}

\tikzset{pics/treeRep/.style n args={1}{code={
            \def\l{1} 
            \foreach \x in {0,1,2,3,4,5}
                \draw (0+\x*\l,0) rectangle (\l+\x*\l,\l) node[pos=0.5]{$\x$};
            \draw [decorate, decoration = {calligraphic brace, mirror,amplitude=10pt}] (0.1,-0.5) --  (5.9,-0.5) node[pos=0.5, below=10pt]{Genome};
            \draw[->] (6.6, 0.5) -- node[above=10pt,align=center, pos=0.5]{translation \\  process} (10.5, 0.5);
            
            \node[rectangle, draw] (head) at (14,2.5) {Condition};
            \node[rectangle, draw] (condition1) at (12.5,0.5) {Condition};
            \node[circle, draw] (leaf1) at (15.5,0.5) {RL};
            \node[circle, draw] (leaf2) at (11.5,-1.5) {RL};
            \node[circle, draw] (leaf3) at (13.5,-1.5) {RL};
            
            \draw[->, shorten <=2pt,shorten >=2pt,>=stealth] (head.south) to[below] node[above]{}(condition1.north) ;
            \draw[->, shorten <=2pt,shorten >=2pt,>=stealth] (head.south) to[below] node[above]{}(leaf1.north) ;
            \draw[->, shorten <=2pt,shorten >=2pt,>=stealth] (condition1.south) to[below] node[above]{}(leaf2.north) ;
            \draw[->, shorten <=2pt,shorten >=2pt,>=stealth] (condition1.south) to[below] node[above]{}(leaf3.north) ;
}}}

\tikzset{pics/gmm/.style n args={0}{code={
        \draw[yscale=0.5,yshift=-3, domain=0:10,blue] plot function{exp(-(x-5)*(x-5)/2/2/2.0)/sqrt(pi)/2};
}}}

\tikzset{pics/vsrRep/.style n args={0}{code={
        \def\l{1}
        \foreach \x in {0,1,2,3,4,5}
            \draw (0+\x*\l,0) rectangle (\l+\x*\l,\l) node[pos=0.5]{$0.\x$};
        \draw [decorate, decoration = {calligraphic brace, mirror,amplitude=10pt}] (0.1,-0.5) --  (5.9,-0.5) node[pos=0.5, below=10pt]{Genome};
        
        \draw[-] (6.6, 0.5) -- node[above=10pt,align=center, pos=0.5]{translation \\  process} (9.5, 0.5);
        \draw[-] (9.5,1.5) -- (9.5,-0.5);
        \draw[->] (9.5,-0.5) -- (10.5,-0.5);
        \draw[->] (9.5,1.5) -- (10.5,1.5);

        \node[draw,xscale=0.1] (BG) at (11.5,1.5) {\vt{\pic{gmm={}};}}; 
        \node[below= 0.1pt of BG,align=center]  {body \\ generator};
        
        \node[draw,xscale=0.1] (BC) at (11.5,-0.5) {\vt{\pic{gmm={}};}}; 
        \node[below= 0.1pt of BC,align=center]  {controller \\ generator};
        
        \node[] (r) at (14.5, 0.5) {\vsr[2mm]{7}{6}{0011110-0011010-1111111-0111110-0100110-0110010}};
        
        \draw[->, shorten <=2pt,shorten >=2pt,>=stealth] (BG.east) to node[]{}(r.west) ;
        \draw[->, shorten <=2pt,shorten >=2pt,>=stealth] (BC.east) to node[]{}(r.west) ;
}}}

\tikzset{pics/direct/.style n args={0}{code={
 \draw [->](-0,0) -- node[above, xshift=5mm, yshift=1mm, align=center] {Model\\complexity} ++ (5, 0);
 \draw (2, -0.25) -- node[below, yshift=-2mm] {Single} ++ (0, 0.5);
 \draw (4, -0.25) -- node[below, yshift=-2mm] {Multiple} ++ (0, 0.5);

 \draw [->](1, 1) -- node[left, yshift=-5mm, align=center] {Task\\complexity} ++ (0,-4);
 \draw (0.75, -1) -- node[right, xshift=2mm] {OpenAI} ++ (0.5, 0);
 \draw (0.75, -2.25) -- node[right, xshift=2mm] {VSRs} ++ (0.5, 0);
}}}

\newcommand{\ndrCycle}[2]{
    \node[circle, draw] (I) at (0*#1,1*#2){I};

    \node[circle, draw] (C) at (1*#1,2*#2){C};
    \node[circle, draw] (B) at (1*#1,1*#2){B};
    \node[circle, draw] (A) at (1*#1,0){A};

    \node[circle, draw] (O) at (2*#1,1*#2){O};

    \draw[->] (I) to[right] node[left]{}(A) ;
    \draw[->] (I) to[right] node[left]{}(B) ;
    \draw[->] (I) to[right] node[left]{}(C) ;

    \draw[<->] (A) to[right] node[left]{}(B) ;
    \draw[<->] (C) to[right] node[left]{}(B) ;

    \draw[->] (A) to[right] node[left]{}(O) ;
    \draw[->] (B) to[right] node[left]{}(O) ;
    \draw[->] (C) to[right] node[left]{}(O) ;
    
}

\newcommand{\ndrCycleHL}[2]{
    \ndrCycle{#1}{#2}
    \draw[red] (1*#1,1.5*#2) ellipse (0.5cm*#1 and 0.8cm*#2);

}

\newcommand{\ndrCycleFK}[2]{
    \node[circle, draw] (I) at (0*#1,1*#2){I};

    \node[circle, draw, red] (F1) at (1*#1,2*#2){F1};
    \node[circle, draw] (A) at (1*#1,0*#2){A};

    \node[circle, draw] (O) at (2*#1,1*#2){O};

    \draw[->] (I) to[right] node[left]{}(F1) ;
    \draw[->] (I) to[right] node[left]{}(A) ;

    \draw[<->] (F1) to[right] node[left]{}(A) ;
    \draw[<->] (A) to[right] node[left]{}(F1) ;

    \draw[->] (F1) to[right] node[left]{}(O) ;
    \draw[->] (A) to[right] node[left]{}(O) ;
}

\newcommand{\ndrCycleFHL}[2]{
    \ndrCycleFK{#1}{#2}
    \draw[blue] (1*#1,1*#2) ellipse (0.5cm*#1 and 1.5cm*#2);
}

\newcommand{\ndrCycleFKp}[2]{
    \node[circle, draw] (I) at (0*#1,1*#2){I};

    \node[circle, draw, blue] (F2) at (1*#1,1*#2){F2};

    \node[circle, draw] (O) at (2*#1,1*#2){O};

    \draw[->] (I) to[right] node[left]{}(F2) ;

    \draw[->] (F2) to[right] node[left]{}(O) ;
}

\tikzset{pics/ndrCyclepc/.style n args={2}{code={
    \ndrCycle{#1}{#2}
}}}

\tikzset{pics/ndrCycleHLpc/.style n args={2}{code={
    \ndrCycleHL{#1}{#2}
}}}
\tikzset{pics/ndrCycleFKpc/.style n args={2}{code={
    \ndrCycleFK{#1}{#2}
}}}
\tikzset{pics/ndrCycleFHLpc/.style n args={2}{code={
    \ndrCycleFHL{#1}{#2}
}}}
\tikzset{pics/ndrCycleFKppc/.style n args={2}{code={
    \ndrCycleFKp{#1}{#2}
}}}

\tikzset{pics/LLT/.style n args={0}{code={
        \node[circle, draw, minimum size=0.7cm] (i0) at (0,0){$x$}; 
        \node[circle, draw, minimum size=0.7cm] (i1) at (0,1){$y$};
        \node[circle, draw, minimum size=0.7cm] (i2) at (0,2){$v_x$};
        \node[circle, draw, minimum size=0.7cm] (i3) at (0,3){$v_y$};
        \node[circle, draw, minimum size=0.7cm] (i4) at (0,4){$\rho$};
        \node[circle, draw, minimum size=0.7cm] (i5) at (0,5){$\Omega$};
        \node[circle, draw, minimum size=0.7cm] (i6) at (0,6){$L_l$};
        \node[circle, draw, minimum size=0.7cm] (i7) at (0,7){$L_r$};
        
        \node[circle, draw, minimum size=0.7cm] (o0) at (4,2){A}; 
        \node[circle, draw, minimum size=0.7cm] (o1) at (4,3){L};
        \node[circle, draw, minimum size=0.7cm] (o2) at (4,4){M};
        \node[circle, draw, minimum size=0.7cm] (o3) at (4,5){R};

        \draw[->, line width=0.5mm] (i0) to[right] node[left]{}(o0) ;
        \draw[->, line width=0.5mm] (i1) to[right] node[left]{}(o0) ;

        \draw[->, line width=0.5mm] (i0) to[right] node[left]{}(o1) ;
        \draw[->, line width=0.5mm] (i2) to[right] node[left]{}(o1) ;
        \draw[->, line width=0.5mm] (i4) to[right] node[left]{}(o1) ;
        \draw[->, line width=0.5mm] (i5) to[right] node[left]{}(o1) ;

        \draw[->, line width=0.5mm] (i1) to[right] node[left]{}(o2) ;
        \draw[->, line width=0.5mm] (i3) to[right] node[left]{}(o2) ;
        \draw[->, line width=0.5mm] (i7) to[right] node[left]{}(o2) ;

        \draw[->, line width=0.5mm] (i4) to[right] node[left]{}(o3) ;
}}}

\tikzset{pics/LLCP/.style n args={0}{code={
        \node[circle, draw, minimum size=0.7cm] (i0) at (0,0){$x$}; 
        \node[circle, draw, minimum size=0.7cm] (i1) at (0,1){$y$};
        \node[circle, draw, minimum size=0.7cm] (i2) at (0,2){$v_x$};
        \node[circle, draw, minimum size=0.7cm] (i3) at (0,3){$v_y$};
        \node[circle, draw, minimum size=0.7cm] (i4) at (0,4){$\rho$};
        \node[circle, draw, minimum size=0.7cm] (i5) at (0,5){$\Omega$};
        \node[circle, draw, minimum size=0.7cm] (i6) at (0,6){$L_l$};
        \node[circle, draw, minimum size=0.7cm] (i7) at (0,7){$L_r$};
        
        \node[circle, draw, minimum size=0.7cm] (o0) at (4,2){A}; 
        \node[circle, draw, minimum size=0.7cm] (o1) at (4,3){L};
        \node[circle, draw, minimum size=0.7cm] (o2) at (4,4){M};
        \node[circle, draw, minimum size=0.7cm] (o3) at (4,5){R};

        \draw[->, line width=0.5mm] (i0) to[right] node[left]{}(o1) ;
        \draw[->, line width=0.5mm] (i2) to[right] node[left]{}(o1) ;
        \draw[->, line width=0.5mm] (i4) to[right] node[left]{}(o1) ;

        \draw[->, line width=0.5mm] (i5) to[right] node[left]{}(o3) ;
        \draw[->, line width=0.5mm] (i2) to[right] node[left]{}(o3) ;
        \draw[->, line width=0.5mm] (i4) to[right] node[left]{}(o3) ;

}}}

\tikzset{pics/MCnhn/.style n args={0}{code={
        \node[circle, draw, minimum size=0.7cm] (i0) at (0,0){$x$}; 
        \node[circle, draw, minimum size=0.7cm] (i1) at (0,2){$v_x$};

        \node[circle, draw, minimum size=0.7cm] (o0) at (4,0){L};
        \node[circle, draw, minimum size=0.7cm](o1) at (4,1){A}; 
        \node[circle, draw, minimum size=0.7cm] (o2) at (4,2){R};

        \draw[->] (i1) to[right] node[left]{}(o0) ;
        \draw[->] (i1) to[right] node[left]{}(o2) ;

}}}

\tikzset{pics/MChn/.style n args={0}{code={
        \node[circle, draw, minimum size=0.7cm] (i0) at (0,0){$x$}; 
        \node[circle, draw, minimum size=0.7cm] (i1) at (0,2){$v_x$};
       
        \node[circle, draw, minimum size=0.7cm] (o0) at (4,0){L};
        \node[circle, draw, minimum size=0.7cm] (o1) at (4,1){A}; 
        \node[circle, draw, minimum size=0.7cm] (o2) at (4,2){R};
        
        \draw[->] (i1) to[right] node[left]{}(o0) ;
        \draw[->] (i1) to[right] node[left]{}(o2) ;
        
        \node[circle, draw, minimum size=0.7cm](h0) at (2,-1){$h_1$};
        \node[circle, draw, minimum size=0.7cm](h1) at (2,3){$h_2$};

        \draw[->] (i0) to[right] node[left]{}(h1) ;
        
        \draw[->] (h1) to[right] node[left]{}(h0) ;
        \draw[->] (h1) to[right] node[left]{}(o0) ;
        \draw[->] (h1) to[right] node[left]{}(o1) ;
        \draw[->] (h1) to[right] node[left]{}(o2) ;

        \draw[->] (h0) to[right] node[left]{}(o0) ;
        \draw[->] (h0) to[right] node[left]{}(o1) ;
        \draw[->] (h0) to[right] node[left]{}(o2) ;

}}}

\tikzset{pics/nettopru/.style n args={3}{code={
        \node[circle, draw, minimum size=0.7cm*#1] (A) at (0*#2,0*#3){A}; 
        \node[circle, draw, minimum size=0.7cm*#1] (B) at (0*#2,2*#3){B};
        \node[circle, draw, minimum size=0.7cm*#1] (C) at (2*#2,1*#3){C};
        \node[circle, draw, minimum size=0.7cm*#1] (D) at (4*#2,0*#3){D}; 
        \node[circle, draw, minimum size=0.7cm*#1] (E) at (4*#2,2*#3){E};

        \draw[-stealth,shorten >=3pt,shorten <=3pt,line width=0.5pt*#1] (A) -- (C);
        \draw[-stealth,shorten >=3pt,shorten <=3pt,line width=0.7pt*#1] (B) -- (C);

        \draw[-stealth,shorten >=3pt,shorten <=3pt,line width=1pt*#1] (C) -- (D);
        \draw[-stealth,shorten >=3pt,shorten <=3pt,line width=0.1pt*#1] (C) -- (E);

}}}

\tikzset{pics/netpru/.style n args={3}{code={
        \node[circle, draw, minimum size=0.7cm*#1] (A) at (0*#2,0*#3){A}; 
        \node[circle, draw, minimum size=0.7cm*#1] (B) at (0*#2,2*#3){B};
        \node[circle, draw, minimum size=0.7cm*#1] (C) at (2*#2,1*#3){C};
        \node[circle, draw, minimum size=0.7cm*#1] (D) at (4*#2,0*#3){D}; 
        \node[circle, draw, minimum size=0.7cm*#1] (E) at (4*#2,2*#3){E};

        \draw[-stealth,shorten >=3pt,shorten <=3pt,line width=0.7pt*#1] (B) -- (C);

        \draw[-stealth,shorten >=3pt,shorten <=3pt,line width=1pt*#1] (C) -- (D);

}}}

%% file: plot-macros.tex
\newcommand{\linesimple}[4]{
    \addplot [each nth point=1,mark=*,mark size=1pt, #4,thick] table [x={#2},y={#3}] {#1};
}

\newcommand{\linewitherror}[5]{
    \addplot [each nth point=20,name path=minuserror,draw=none,no markers,forget plot] table [x={#2},y expr=\thisrow{#3}-\thisrow{#4}] {#1};
    \addplot [each nth point=20,name path=pluserror,draw=none,no markers,forget plot] table [x={#2},y expr=\thisrow{#3}+\thisrow{#4}] {#1};
    \addplot [forget plot,fill=#5,opacity=0.1] fill between[on layer={},of=pluserror and minuserror];
    \addplot [each nth point=20,#5,thick,no markers,legend image code/.code={\fill [fill=#5, opacity=0.1, draw=none] (0mm,-1ex) -- (0mm,1ex) -- (6mm,1ex) -- (6mm,-1ex) -- cycle; \draw [#5,thick] (0mm,0mm) -- (6mm,0mm);}] table [x={#2},y={#3}] {#1};
}

\newcommand{\linewitherrorns}[5]{
    \addplot [name path=minuserror,draw=none,no markers,forget plot] table [x={#2},y expr=\thisrow{#3}-\thisrow{#4}] {#1};
    \addplot [name path=pluserror,draw=none,no markers,forget plot] table [x={#2},y expr=\thisrow{#3}+\thisrow{#4}] {#1};
    \addplot [forget plot,fill=#5,opacity=0.1] fill between[on layer={},of=pluserror and minuserror];
    \addplot [#5,thick,mark=*,mark size=1pt,legend image code/.code={\fill [fill=#5, opacity=0.1, draw=none] (0mm,-1ex) -- (0mm,1ex) -- (6mm,1ex) -- (6mm,-1ex) -- cycle; \draw [#5,thick] (0mm,0mm) -- (6mm,0mm);}] table [x={#2},y={#3}] {#1};
}

\newcommand{\pvalue}[5]{
    \draw (#1,#3) -- (#1,#3-#4);
    \draw (#2,#3) -- (#2,#3-#4);
    \draw (#1,#3) -- node [above,scale=.65] {#5} (#2,#3);
}

\def\addlegendimage{\csname pgfplots@addlegendimage\endcsname}

\definecolor{cola1}{RGB}{228,26,28}
\definecolor{cola2}{RGB}{55,126,184}
\definecolor{cola3}{RGB}{77,175,74}
\definecolor{cola4}{RGB}{152,78,163}
\definecolor{cola5}{RGB}{255,127,0}
\definecolor{cola6}{RGB}{255,255,51}
\definecolor{cola7}{RGB}{166,86,40}
\definecolor{cola8}{RGB}{176,13,13}
\definecolor{cola9}{RGB}{8,92,156}

\definecolor{colb1}{RGB}{102,194,165}
\definecolor{colb2}{RGB}{252,141,98}
\definecolor{colb3}{RGB}{141,160,203}
\definecolor{colb4}{RGB}{231,138,195}

\definecolor{colb1}{RGB}{102,194,165}
\definecolor{colb2}{RGB}{252,141,98}
\definecolor{colb3}{RGB}{141,160,203}
\definecolor{colb4}{RGB}{231,138,195}

\pgfplotsset{/pgfplots/colormap={evolvability}{rgb255=(241,163,64) rgb255=(224,224,224) rgb255=(153,142,195)}}
\pgfplotsset{/pgfplots/colormap={fitness}{rgb255=(229,245,249) rgb255=(153,216,201) rgb255=(44,162,95)}}
\pgfkeys{/pgf/number format/1000 sep={\,}}